\PassOptionsToPackage{table}{xcolor}
\documentclass[10pt,twocolumn,letterpaper]{article}

\usepackage[pagenumbers]{cvpr}

\usepackage{booktabs}

\usepackage[table]{xcolor}
\newsavebox{\fitbox}
\newcommand{\fitto}[2]{\sbox{\fitbox}{#2}\ifdim\wd\fitbox>#1\resizebox{#1}{!}{\usebox{\fitbox}}\else\usebox{\fitbox}\fi}
\newcommand{\tnmark}[1]{\rlap{$^{#1}$}}
\newcommand{\unr}[1]{\textcolor{gray}{#1}} %
\definecolor{linkblue}{rgb}{0.21,0.49,0.74}
\usepackage[pagebackref,breaklinks,colorlinks,allcolors=linkblue]{hyperref}

\newcommand{\ourmethod}{\textit{AnchorMan}}

\title{Anchor the Man, Move the Camera}

\author{Yuntian Ke$^{1}$\quad Weijie Lyu$^{2}$\quad Shuo Chen$^{2}$\quad Zhixin Shu$^{3}$\quad Ming-Hsuan Yang$^{2}$\\
$^{1}$University of Pennsylvania\quad $^{2}$University of California, Merced\quad $^{3}$Adobe Research
}

\begin{document}
\captionsetup{hypcap=false}
\twocolumn[{%
\renewcommand\twocolumn[1][]{#1}%
\maketitle
\begin{center}
\vspace{-4mm}
\centering
\includegraphics[width=0.98\linewidth]{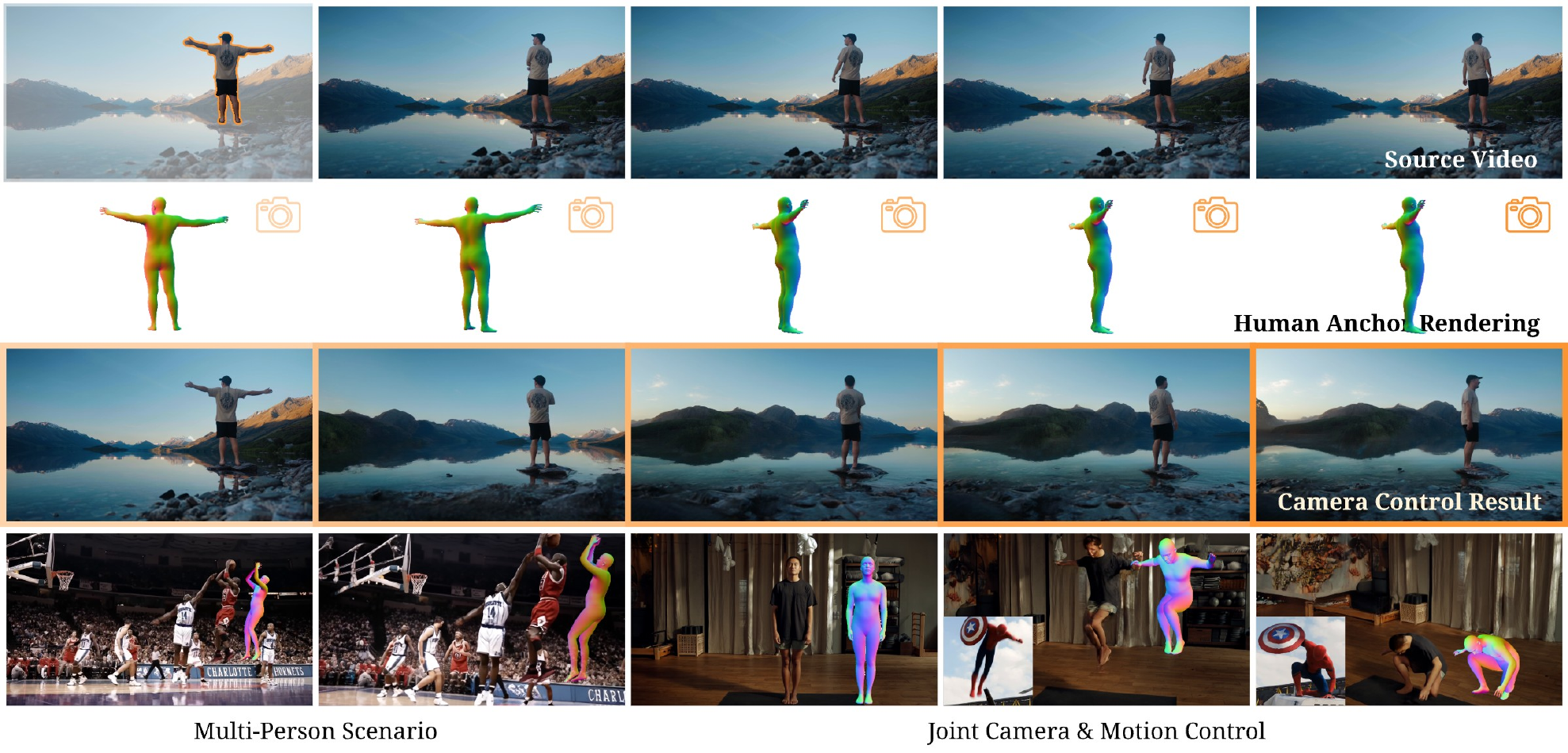}
\captionof{figure}{{\ourmethod} uses the reconstruction of a human in the scene as the anchor to control the camera for video generation. It can be applied to multi-person scenarios. Given an image input and a driving video, the anchor representation can also be used for motion control.}
\label{fig:teaser}
\end{center}
}]

\begin{abstract}
Existing camera-controlled video generation methods often fail when a person is the main subject.
Due to the ambiguous implicit camera representation or inaccurate explicit geometry estimation, the generated video often displays incoherent camera changes between the human and the background, as well as distorted human structure or motion.
We introduce \ourmethod{}, a video-to-video method that uses the human body as a geometric anchor for camera control.
Given a source video and a target camera trajectory, we reconstruct a body mesh from the first frame and render it along the estimated source trajectory and the requested target trajectory.
The paired renderings specify the viewpoint change relative to the subject, independent of the unknown scene scale, and supply body geometry for sides the source never shows.
Replacing the fixed mesh with a sequence of posed meshes extends the same condition to motion control.
We train a bidirectional model on synthetic renderings and real captures, and distill a causal variant that accepts updated camera commands as it generates successive chunks.
On an extensive camera-control benchmark, \ourmethod{} achieves the highest joint success rate across camera, subject, and image-quality checks and the lowest body-orientation error on orbit trajectories.
We plan to release the model weights and source code. Project page: \url{https://kytttt.github.io/AnchorMan/}.

\end{abstract}

\begin{figure*}[t]
\centering
\includegraphics[width=\textwidth]{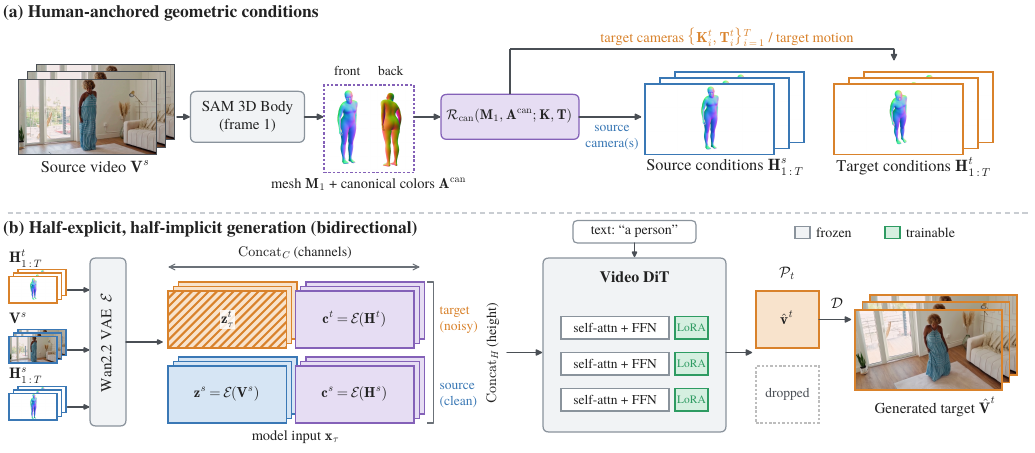}
\caption{\textbf{Overview of {\ourmethod}.} (a) For camera control, SAM 3D Body reconstructs a mesh $\mathbf{M}_1$ from the first source frame. This frozen mesh is rendered as canonical normal maps under the estimated source cameras and the target cameras to obtain $\mathbf{H}^{s}$ and $\mathbf{H}^{t}$; subject motion comes from the source video. (b) Clean source and noisy target latents are concatenated along height, and the encoded canonical normal maps are concatenated along channels. The model infers the relative camera motion from $\mathbf{H}^{s}$ and $\mathbf{H}^{t}$ and draws scene context from the source video.}
\label{fig:overview}
\end{figure*}

\section{Introduction}
\label{sec:intro}
Camera-controlled video-to-video generation reshoots a source video along a new camera trajectory while preserving its content and motion.
Human-centric videos are a major application of this task, with uses in film production, sports and performance replays, and social media content creation.
In such cases, both the person and the background must change consistently with the requested camera motion, and the person's appearance must remain plausible from viewpoints the source may never show.

Existing approaches commonly condition generation on implicit camera parameters or explicit geometry projected into the target camera poses.
The former~\cite{he2024cameractrl, bai2025recammaster} suffer from scale ambiguity: the parameters do not specify how far the person is from the camera, yet the same camera translation moves a nearby person much farther across the image than a distant one, and also changes their viewing angle more.
The generated person therefore often appears in the wrong position or scale.
The latter, such as TrajectoryCrafter~\cite{yu2025trajectorycrafter} and GEN3C~\cite{ren2025gen3c}, resolve this ambiguity with estimated depth but inherit its errors on the human body.
A frontal view provides little geometry for the back of a person, so large rotations expose missing or implausible surfaces; depth errors on a moving limb stretch or fragment its projection, so limbs appear distorted or disconnected.
Neither representation gives the generator reliable geometry for the person, so the background often follows the requested camera while the view of the person changes too little, staying close to their source view.
This motivates a camera representation designed specifically for human subjects.

When filming a person, a shot is naturally described relative to the subject: the side from which they are seen, the camera's distance, and the size and position of the person in the frame.
Following this intuition, we propose \ourmethod{}, a camera-controlled video generation method that represents the camera through the person, using the human body as a geometric anchor.
Given a source video and a target camera trajectory, we reconstruct a body mesh from the first frame, fix its pose, and render it under both the source cameras estimated from the video and the requested target cameras (\cref{fig:overview}).
The two renderings thus differ only in viewpoint and encode the requested camera change, leaving the person's motion to be taken from the source video.
This human anchor addresses both limitations of existing representations.
First, the rendering removes the scale ambiguity: since the trajectory is defined relative to the person, it shows where the person appears in the target view and how large they appear, regardless of the unknown scene scale.
Second, the body mesh provides complete geometric guidance for the person: it covers sides of the body that the source never shows, and it avoids the stretched or fragmented limbs that depth errors produce in projected point clouds.
The representation is thus half explicit and half implicit: the mesh makes the subject's geometry explicit, while the generator infers the surrounding scene from source appearance.

The human anchor also extends to motion control.
Instead of freezing the mesh, we reconstruct a mesh from each frame of a driving video and render the sequence under a fixed camera, so the same model animates a reference image without an additional motion module.
Unlike animation methods that condition on body cues for motion alone~\cite{hu2023animateanyone, zhu2024champ, cheng2025wananimate}, \ourmethod{} uses one shared representation for both camera and motion control.

The bidirectional model denoises the full clip at once and therefore requires the entire trajectory in advance.
For interactive use, where camera commands arrive during generation, we distill a causal variant with distribution matching distillation (DMD)~\cite{dmd}.
It generates one chunk at a time in four denoising steps, accepts updated camera conditions for each new chunk, and keeps a bounded attention cache for streaming inference.
Our contributions are as follows:
\begin{itemize}
\item We propose \ourmethod{}, which conditions camera-controlled video generation on a frozen body mesh rendered along the source and target trajectories, resolving the scale ambiguity and incomplete body geometry of existing camera representations.
\item By replacing the frozen mesh with per-frame meshes from a driving video, we show that the same condition supports both camera and motion control in one model.
\item We distill a causal variant that generates one chunk at a time in four denoising steps and accepts updated camera commands during generation.
\item We evaluate with a joint success criterion that requires camera, subject, and image-quality checks to pass together, where \ourmethod{} achieves the highest joint success rate and the lowest body-orientation error among all compared methods.
\end{itemize}

\section{Related work}
\label{sec:related}

\vspace{1mm}
\noindent\textbf{Camera-controlled video generation.}
Camera control in video diffusion models is commonly introduced through camera parameters, injected as extrinsics~\cite{wang2024motionctrl} or Pl\"ucker rays~\cite{he2024cameractrl,bahmani2025ac3d,he2025cameractrl2}.
For video inputs, re-capture methods render the source content along a new trajectory.
Some condition implicitly on the relative camera pose or camera-aware positional encodings~\cite{vanhoorick2024gcd,bai2025recammaster,park2025redirector,park2026flashrender}, optionally aligned with metric geometry rewards~\cite{li2026geoalign}.
Others project estimated geometry into the target view, through depth warps and point clouds~\cite{zhang2025recapture,yu2024viewcrafter,yu2025trajectorycrafter,ren2025gen3c} or meshes reconstructed from the source video~\cite{hu2025ex4d,chen2026depthdirector}, and feed-forward 4D reconstruction has also been used to re-render dynamic scenes~\cite{yang2026neoverse}.

\vspace{1mm}
\noindent\textbf{Human video generation.}
Pose-guided methods animate a reference image with a sequence of body conditions, ranging from 2D skeletons~\cite{hu2023animateanyone,zhang2024mimicmotion,tu2025stableanimator,cheng2025wananimate} to DensePose~\cite{xu2024magicanimate} and rendered SMPL depth, normal, and semantic maps~\cite{zhu2024champ,zhou2025realisdance}.
HumanVid~\cite{wang2024humanvid} extends this setting to camera-controllable animation by conditioning on both body pose and camera trajectory, and other methods control camera and motion jointly through rendered 3D proxies, aligning SMPL-X characters with scene point clouds~\cite{cao2025uni3c} or rendering 3D tracking videos~\cite{gu2025das}.
Uni3C~\cite{cao2025uni3c}, the closest to our setting, animates a single reference image and builds its scene condition from monocular depth, whereas \ourmethod{} reshoots a source video and requires no scene geometry.
A related line synthesizes novel views of a person from video, using SMPL-conditioned 4D transformers~\cite{shaohuman4dit}, normal-map-conditioned multi-view generation~\cite{zhi2025mvperformer}, or skeleton-conditioned 4D generation~\cite{jin4danyone}, while portrait videos can be re-rendered along new camera trajectories~\cite{lyu2026facecam}.

\vspace{1mm}
\noindent\textbf{Autoregressive video generation and distillation.}
Causal video diffusion generates videos sequentially when future conditions are unavailable, and Diffusion Forcing~\cite{chen2024diffusionforcing} enables this by training with independent per-token noise levels.
CausVid~\cite{yin2025causvid} distills a bidirectional teacher into a few-step causal student with distribution matching distillation~\cite{dmd,yin2024dmd2}.
Self Forcing~\cite{huang2025selfforcing} trains on the student's own rollouts, Causal Forcing~\cite{zhu2026causalforcing} studies teacher--student causal compatibility, and later methods extend such students to long, real-time streaming generation~\cite{liu2026rollingforcing,yang2026longlive}.
Interactive world models further condition autoregressive generation on user actions or camera commands~\cite{he2025matrixgame2,inspatio2026world}.

In contrast, \ourmethod{} conditions video-to-video generation on a body mesh rendered along both the source and target trajectories, a design that supports camera control, motion control, and causal generation.

\section{Method}
\label{sec:method}
\subsection{Problem formulation}
\label{sec:method:formulation}
Given a source video $\mathbf{V}^s = \{\mathbf{I}^s_i\}_{i=1}^T$ of a person and a target camera trajectory, \ourmethod{} generates a target video that follows the trajectory while preserving the subject's appearance and the source context.
We reconstruct a body mesh from the first source frame and render it with canonical normal colors under the source and target cameras, which gives the condition sequences $\mathbf{H}^s$ and $\mathbf{H}^t$ (\cref{sec:method:anchor}).
A video diffusion model then generates
\begin{equation}
\hat{\mathbf{V}}^t = F_\theta(\mathbf{V}^s, \mathbf{H}^s, \mathbf{H}^t),
\label{eq:formulation}
\end{equation}
where $F_\theta$ includes VAE encoding and the fixed text prompt \texttt{a person} (\cref{sec:method:generation}).
For camera control, the fixed mesh is rendered under changing cameras, and the source can also be a single image repeated over time.
For motion control, the source is a reference image $\mathbf{I}_{\text{ref}}$ repeated as $\mathbf{V}^s = \{\mathbf{I}_{\text{ref}}\}_{i=1}^T$, and $\mathbf{H}^t$ renders per-frame meshes reconstructed from a driving video (\cref{sec:method:motion}).
Besides this bidirectional model, we distill a causal variant with DMD~\cite{dmd} that generates four-step chunks and accepts updated camera commands during generation (\cref{sec:method:autoregressive}).

\subsection{Human-anchored geometric conditioning}
\label{sec:method:anchor}
\ourmethod{} represents the target camera trajectory through renderings of a frozen human mesh.
Each rendered map specifies where the subject appears in the image, how large they appear, and which side of the body faces the camera.
The generation model is trained to infer human motion and the surrounding scene from the source video.

We estimate the body mesh with SAM 3D Body~\cite{yang2026sam3dbody}, whose parametric mesh provides a complete surface, including a back side inferred from the body prior.
To give these renderings a consistent surface coordinate system, we compute vertex normals on a shared Momentum Human Rig (MHR) template in its canonical rest pose, map each normal $\mathbf{n}$ to RGB as $255(\mathbf{n}+1)/2$, and transfer these colors by vertex index.
Each vertex keeps its color as pose and viewpoint change, so the rendering directly shows which side of the body faces the camera.
At frame $t$, the condition is
\begin{equation}
\mathbf{H}_t = \mathcal{R}_{\mathrm{can}}\big(\mathbf{M}_t, \mathbf{A}^{\mathrm{can}}; \mathbf{K}_t, \mathbf{T}_t\big),
\label{eq:condition}
\end{equation}
where $\mathbf{M}_t$ is the mesh, $\mathbf{A}^{\mathrm{can}}$ contains its canonical colors, and $\mathbf{K}_t$ and $\mathbf{T}_t$ are the rendering intrinsics and the mesh-to-camera transformation.

At inference, we reconstruct $\mathbf{M}_1$ from the first source frame and freeze its pose and shape.
ViPE~\cite{huang2025vipe} estimates per-clip intrinsics and per-frame source cameras, and we place $\mathbf{M}_1$ at the median ViPE depth within the person mask.
Because our trajectories are defined relative to the subject, the overall scene scale does not affect the conditions.
Rendering $\mathbf{M}_1$ under the source and target cameras gives $\mathbf{H}^s$ and $\mathbf{H}^t$, whose difference encodes the relative camera motion.
For a static source camera, $\mathbf{H}^s$ reduces to a repeated first-frame rendering.

\subsection{Extension to motion control}
\label{sec:method:motion}
The same human-anchored condition also supports motion control.
Given a reference image and a driving video, we reconstruct a mesh from the reference image and render it as the source condition, and reconstruct a mesh from each driving frame and render it with the same canonical colors as the target condition.
Unlike camera control, (1) the source input is the reference image repeated over time, conveying identity and background information; (2) the target condition renders a mesh sequence that moves with the driving video rather than a frozen mesh under changing cameras, so the human motion comes from the driving video. For joint camera and motion control, the per-frame meshes are rendered under moving target cameras.

\subsection{Model and training}
\label{sec:method:generation}
\noindent\textbf{Architecture.}
We adapt Wan2.2-Fun-5B-Control~\cite{alibabapai2025wanfun} with LoRA while freezing its pretrained DiT weights.
Let $\mathbf{z}^s$ and $\mathbf{z}^t_\tau$ denote the clean source and noisy target video latents, and let $\mathbf{c}^s$ and $\mathbf{c}^t$ denote their encoded condition maps.
We concatenate the latents along the height dimension and attach the condition maps along the channel dimension:
\begin{equation}
\mathbf{z}_{\tau}
{=} \operatorname{Con}_{H}
(\mathbf{z}^{t}_{\tau}, \mathbf{z}^{s}),
\mathbf{c}
{=} \operatorname{Con}_{H}
(\mathbf{c}^{t}, \mathbf{c}^{s}),
\mathbf{x}_{\tau}
{=} \operatorname{Con}_{C}
(\mathbf{z}_{\tau}, \mathbf{c}).
\label{eq:spatial_conditioning}
\end{equation}
Full self-attention over $\mathbf{x}_\tau$ lets target tokens attend directly to the clean source latents.
Placing the target above the source lets the target use the same height RoPE indices as in single-stream generation, and keeps corresponding source and target frames at the same temporal position for chunk-wise causal attention (\cref{sec:method:autoregressive}).
Source latents are never noised. All tokens share the sampled timestep embedding.

\vspace{1mm}
\noindent\textbf{Objective.}
We supervise the target half with weighted flow matching:
\begin{equation}
\mathcal{L}_{\mathrm{FM}} = \mathbb{E}_{\tau,\boldsymbol{\epsilon}}\Big[ w(\tau) \big\| \mathcal{P}_t\big(f_\theta(\mathbf{x}_\tau, \tau)\big) - \mathbf{v}^t_\tau \big\|_2^2 \Big],
\label{eq:target_flow_matching}
\end{equation}
where $f_\theta$ predicts velocity, $\mathcal{P}_t$ selects the target half, and $w(\tau)$ is the scheduler's timestep-dependent weight.
With Gaussian noise $\boldsymbol{\epsilon}$ and $\tau$ sampled uniformly over the discretized schedule, the noisy target is $\mathbf{z}^t_\tau = (1-\sigma_\tau)\mathbf{z}^t_0 + \sigma_\tau \boldsymbol{\epsilon}$ and its velocity is $\mathbf{v}^t_\tau = \boldsymbol{\epsilon} - \mathbf{z}^t_0$.

\begin{figure}[t]
\centering
\includegraphics[width=\columnwidth]{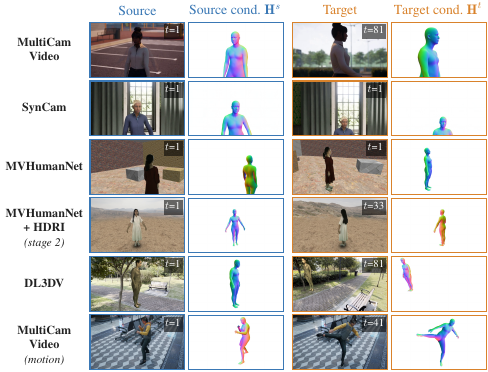}
    \caption{\textbf{Examples of constructed training pairs.} Rows show MultiCamVideo~\cite{bai2025recammaster}, SynCam~\cite{bai2025syncammaster}, MVHumanNet~\cite{xiong2024mvhumannet} with rendered scenes, and DL3DV~\cite{ling2024dl3dv} with an inserted textured human. Columns show a source frame, its canonical condition, a target frame, and the corresponding target condition. The final row shows a MultiCamVideo motion-control pair.}
\label{fig:data_sources}
\end{figure}

\subsection{Training data construction}
\label{sec:method:data}

We construct paired source and target videos with canonical human conditions from four complementary data sources (\cref{fig:data_sources}).
Camera-control pairs render a fixed reference mesh along the source and target camera paths, while motion-control pairs use per-frame target meshes.
This construction combines captured human motion with scene context and camera trajectories, providing examples of viewpoint changes, human motion, and their joint variation. Per-source construction details are given in \cref{sec:supp:data_construction}.

\subsection{Autoregressive generation}
\label{sec:method:autoregressive}

\begin{figure}[t] \centering \includegraphics[width=\columnwidth]{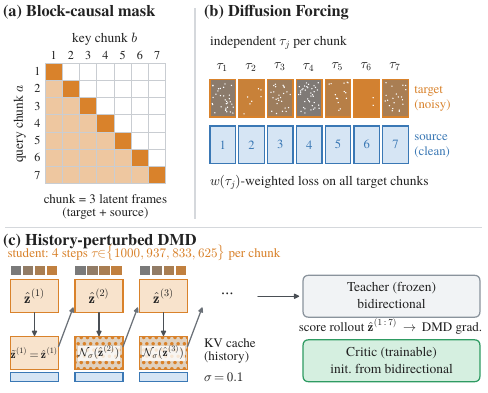} \caption{\textbf{Causal generation and distillation.} (a) Block-causal attention permits access to current and preceding chunks. (b) Diffusion Forcing independently corrupts target chunks. (c) The DF-initialized student generates 7 chunks with 4 denoising steps each. The first generated chunk is cached without perturbation; later chunks receive noise when reused as history.} \label{fig:causal_dmd} \end{figure}

Starting from the initial bidirectional checkpoint, we add a block-causal attention mask (\cref{fig:causal_dmd}a): attention is bidirectional within a chunk, and chunk $a$ attends to chunk $b$ only if $b\leq a$. Each chunk accesses current and preceding source frames and past target chunks. Future geometric conditions can therefore be updated without regenerating completed chunks.

\vspace{1mm}
\noindent\textbf{Diffusion Forcing.}
We train on camera-control pairs with Diffusion Forcing (DF)~\cite{chen2024diffusionforcing}, independently sampling a noise timestep for every target chunk, including the first (\cref{fig:causal_dmd}b). All target chunks are jointly supervised using \cref{eq:target_flow_matching}, while source latents remain clean. This exposes the model to noisy target history. In our preliminary trials with teacher forcing~\cite{williams1989learning}, clean ground-truth history appeared to encourage copying and weaken newly visible background synthesis during self-rollout.

\vspace{1mm}
\noindent\textbf{History-perturbed DMD.}
We initialize the causal student from the DF checkpoint, and the frozen teacher and trainable critic from the initial bidirectional checkpoint. The teacher and critic process full rollouts bidirectionally, while the student generates causally in four denoising steps per chunk (\cref{fig:causal_dmd}c).

During both DMD training and inference, we perturb generated target chunks when caching them as history, except for the first chunk:
\begin{equation}
\tilde{\mathbf{z}}^{(j)} =
\begin{cases}
\hat{\mathbf{z}}^{(j)}, & j = 1,\\
\mathcal{N}_{\sigma}
(\hat{\mathbf{z}}^{(j)}), & j > 1,
\end{cases}
\label{eq:history_noise}
\end{equation}
where $\hat{\mathbf{z}}^{(j)}$ is the generated latent of chunk $j$ and $\mathcal{N}_{\sigma}(\hat{\mathbf{z}})=(1-\sigma)\hat{\mathbf{z}}+\sigma\boldsymbol{\epsilon}$, with $\boldsymbol{\epsilon}\sim\mathcal{N}(\mathbf{0},\mathbf{I})$. We use $\sigma=0.1$ and condition history tokens on their noise level.

\vspace{1mm}
\noindent\textbf{Long-video inference.}
Beyond the 81-frame training length, we bound the KV cache: each chunk attends to itself, the first chunk, and the six most recent chunks, and cached keys keep their absolute temporal positions.
The first chunk serves as an attention sink~\cite{xiao2024streamingllm} and is cached without perturbation, as in \cref{eq:history_noise}; an 81-frame output fits entirely within this window.

\section{Experiments}
\label{sec:experiments}

We evaluate \ourmethod{} on camera control in three settings: in-the-wild human videos without ground-truth target views; multi-view captured videos with ground-truth target views; and background-free videos and single images for full 360$^\circ$ orbit generation compared against camera control and human reconstruction methods. We also evaluate motion control and joint camera and motion control.

\vspace{1mm}
\noindent\textbf{Implementation.}
We train the bidirectional model for 30k steps on multi-view human and scene data and distill a causal variant with Diffusion Forcing and DMD; data mixtures and schedules are given in \cref{sec:supp:additional_details}. Outputs have 81 frames at $832\times480$ with the prompt \texttt{a person}, and bidirectional sampling uses 50 steps.

\subsection{Camera control}
\label{sec:experiments:camera}
\vspace{1mm}
\noindent\textbf{Protocol.}
We evaluate 580 source--trajectory pairs: 40 videos with 11 trajectories each (\emph{Standard}) and 20 videos with 7 harder trajectories each (\emph{Extended}),
drawn from OpenVid~\cite{nan2024openvid}, Vista4D~\cite{lin2026vista4d}, DAVIS~\cite{ponttuset2017davis}, Redirect4D-Bench~\cite{cao2026redirect4d},
and HumanVid~\cite{wang2024humanvid}. Trajectories include pans, tilts, zooms, orbits with and without return, roll, and lateral truck.
We compare with ReCamMaster~\cite{bai2025recammaster}, TrajectoryCrafter~\cite{yu2025trajectorycrafter}, ReDirector~\cite{park2025redirector}, GEN3C~\cite{ren2025gen3c},
DepthDirector~\cite{chen2026depthdirector}, Geo-Align~\cite{li2026geoalign}, FlashRender~\cite{park2026flashrender}, InSpatio-World~\cite{inspatio2026world},
and NeoVerse~\cite{yang2026neoverse}, all with the same input crop and camera trajectory (see \cref{sec:camera_eval_details}).

\vspace{1mm}
\noindent\textbf{Metrics.}
Visual quality uses MEt3R~\cite{asim2025met3r}, five VBench~\cite{huang2023vbench} dimensions, and FID/FVD against the source videos. Subject-level accuracy uses GVHMR~\cite{shen2024gvhmr} body-orientation error (SAM 3D Body~\cite{yang2026sam3dbody} and 4D-Humans~\cite{goel2023humans4d} give consistent rankings), shoulder-midpoint position error normalized by image height $H$, and the rate of frames in which the expected subject is missing; motion preservation compares the root-aligned GVHMR body pose of the source and generated videos. A case reaches joint success only when all applicable camera, orientation, person, and image-clarity checks pass, with thresholds set relative to no-motion and no-turn references (see \cref{sec:joint_sensitivity}).

\begin{figure*}[t]
\centering
\includegraphics[width=0.98\linewidth]{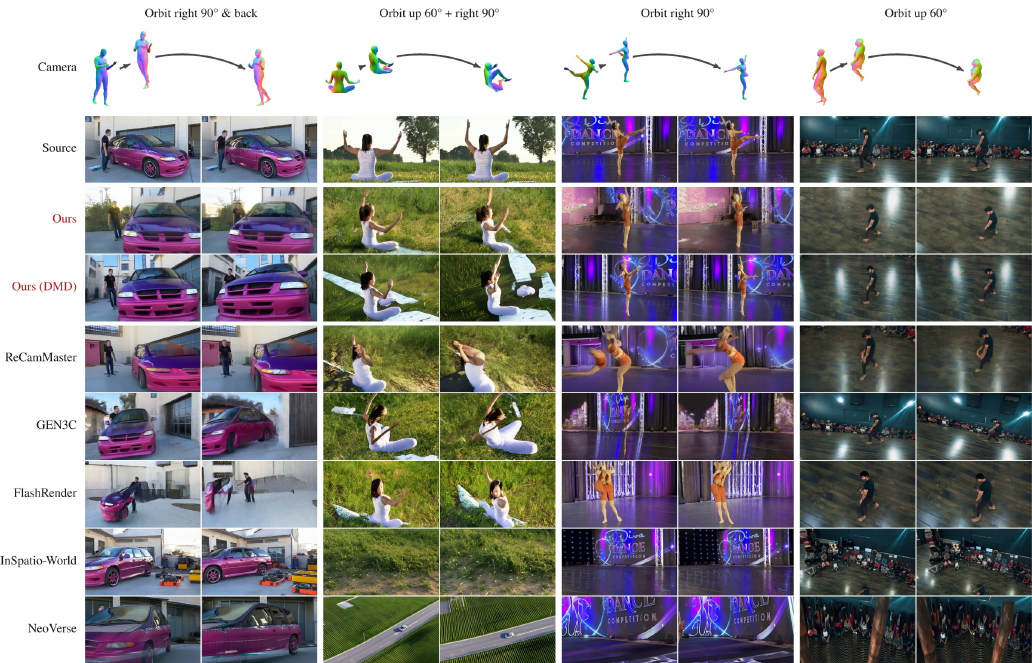}
\caption{\textbf{Qualitative comparison on camera control.}
\ourmethod{} places the person correctly, shows the requested side of the body, and preserves body proportions, while baselines misplace, misorient, or distort the person.
Four baselines are omitted for space; all methods are shown in \cref{fig:qual:camera_all}.}
\label{fig:qual:camera}
\end{figure*}

\vspace{1mm}
\noindent\textbf{Results.}
Background camera accuracy separates the methods far less than human orientation does (\cref{tab:exp:camera}):
among non-autoregressive methods, mean ViPE rotation error spans $2.10^\circ$--$10.65^\circ$, whereas body-orientation error on orbit cases
spans $12.6^\circ$--$39.9^\circ$. \ourmethod{} (bidirectional) is first in orientation,
with the best position and size accuracy. Under a source-clustered paired bootstrap, \ourmethod{} has significantly lower orientation error than every compared method.
Typical failures either leave the person at a wrong orientation as the scene moves
or distort the body under large viewpoint changes (\cref{fig:qual:camera}).

\begin{figure}[t]
\centering
\includegraphics[width=\linewidth]{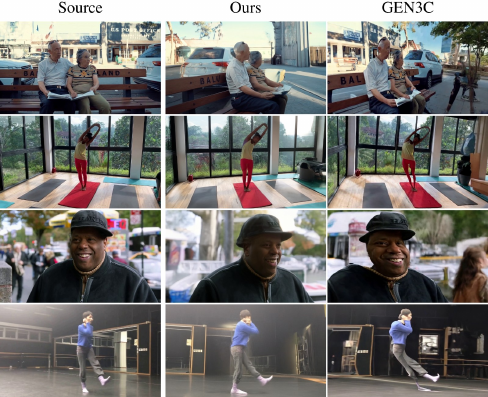}
\caption{\textbf{Cases that both GEN3C and \ourmethod{} pass under the joint evaluation.} A plausible overall framing can still hide distorted proportions:
compare the body shape in the paired outputs. More cases are shown in \cref{fig:gen3c_pass_more}.}
\label{fig:gen3c_pass}
\end{figure}

\ourmethod{} reaches the highest joint success and ranks first in 10 of 11 threshold settings
(\cref{tab:exp:camera,tab:sens:Standard_Extended}).
The joint criterion, however, checks only where the person is placed and which way the body faces, not body proportions:
pose estimators fit a plausible skeleton even to a flattened or twisted body.
As a result, GEN3C, the closest baseline under this criterion, often passes the checks while distorting the body, as seen in the qualitative comparison (\cref{fig:qual:camera}) and in cases that both methods pass (\cref{fig:gen3c_pass}), whereas \ourmethod{} preserves the body's proportions.
Consistent with this, among cases where both methods turn the body correctly, GEN3C's estimated pose departs further from the source motion than that of \ourmethod{}: after subtracting the pose estimator's own viewpoint noise, its error is 8.41\,mm higher (95\% CI [3.85, 13.18]; \cref{sec:supp:motion_preservation}).
On multi-view captures, where silhouettes can be compared with ground-truth views, GEN3C also has lower person IoU than \ourmethod{} (\cref{tab:exp:nvs}).

\begin{table*}[!t]
\centering
\caption{\textbf{Background-level and subject-level camera control on in-the-wild human videos.} Joint is the joint success rate over all cases. Rows below the rule are autoregressive. Orient.\ is the GVHMR body-orientation error ($^\circ$) on orbit cases, and F/S/B (front/side/back) accuracy uses GVHMR. Size is the absolute log ratio of observed to expected box height on eligible full-body frames. MPJPE and PA-MPJPE (mm) compare the root-aligned body pose of the source and the generated video on orbit trajectories (GVHMR); they also decrease when the view is not rotated, so they should be read together with the orientation error.}
\label{tab:exp:camera}
\small\setlength{\tabcolsep}{3.5pt}\renewcommand{\arraystretch}{0.88}\setlength{\aboverulesep}{0.8pt}\setlength{\belowrulesep}{0.8pt}\setlength{\cmidrulesep}{0pt}
\fitto{\textwidth}{%
\begin{tabular}{lcccccccccc}\toprule
& & \multicolumn{2}{c}{Background (ViPE)} & \multicolumn{2}{c}{Human (GVHMR)} & \multicolumn{3}{c}{Subject placement} & \multicolumn{2}{c}{Motion (mm)} \\
\cmidrule(lr){3-4}\cmidrule(lr){5-6}\cmidrule(lr){7-9}\cmidrule(lr){10-11}
Method & Joint (\%)$\uparrow$ & Rot.$\downarrow$ & Trans.$\downarrow$ & Orient.$\downarrow$ & F/S/B (\%)$\uparrow$ & Pos.\ ($/H$)$\downarrow$ & Miss (\%)$\downarrow$ & Size err.$\downarrow$ & MPJPE$\downarrow$ & PA$\downarrow$ \\
\midrule
ReCamMaster~\cite{bai2025recammaster} & 42.24 & 9.96 & 0.048 & 19.71 & 77.01 & 0.148 & \textbf{2.8} & 0.123 & \textbf{55.57} & \underline{45.12} \\
TrajectoryCrafter~\cite{yu2025trajectorycrafter} & 40.34 & 7.51 & 0.056 & 39.85 & 64.51 & 0.227 & 15.1 & 0.224 & 104.31 & 87.18 \\
ReDirector~\cite{park2025redirector} & 55.52 & 6.77 & 0.037 & 19.82 & 80.99 & 0.153 & 7.9 & 0.139 & 69.42 & 55.79 \\
GEN3C~\cite{ren2025gen3c} & \underline{75.52} & \underline{3.19} & \textbf{0.019} & 20.52 & 80.47 & \underline{0.062} & 3.6 & \underline{0.094} & 65.53 & 52.62 \\
DepthDirector~\cite{chen2026depthdirector} & 48.10 & 10.65 & 0.058 & 28.84 & 73.48 & 0.185 & 17.6 & 0.169 & 71.43 & 57.53 \\
Geo-Align~\cite{li2026geoalign} & 55.00 & 6.15 & 0.035 & 21.33 & 78.92 & 0.156 & 6.9 & 0.124 & 68.68 & 55.49 \\
FlashRender~\cite{park2026flashrender} & 62.07 & 7.28 & 0.037 & \underline{16.99} & \underline{82.36} & 0.125 & 4.6 & 0.102 & 62.86 & 50.13 \\
NeoVerse~\cite{yang2026neoverse} & 52.24 & \textbf{2.10} & \underline{0.020} & 35.27 & 70.76 & 0.159 & 8.2 & 0.185 & 86.07 & 70.19 \\
\ourmethod{} & \textbf{78.62} & 7.16 & 0.030 & \textbf{12.61} & \textbf{87.25} & \textbf{0.059} & \underline{3.3} & \textbf{0.084} & \underline{55.66} & \textbf{44.84} \\
\midrule
InSpatio-World~\cite{inspatio2026world} & 58.10 & \textbf{4.76} & \textbf{0.042} & 23.50 & \textbf{77.75} & 0.154 & 8.0 & 0.219 & 71.44 & 56.76 \\
\ourmethod{}-DMD & \textbf{65.52} & 9.29 & 0.048 & \textbf{20.25} & 75.78 & \textbf{0.062} & \textbf{1.9} & \textbf{0.091} & \textbf{49.56} & \textbf{39.73} \\
\bottomrule\end{tabular}}
\end{table*}

\ourmethod{}-DMD has the best visual quality among all methods (\cref{tab:supp:quality}): it ranks first on MEt3R, FID, the VBench mean,
and four VBench dimensions and second on FVD.
The bidirectional model scores slightly lower on these metrics but has stronger orientation control. Within the autoregressive group, \ourmethod{}-DMD exceeds InSpatio-World on joint success in every split. On an RTX 6000 Ada, \ourmethod{}-DMD generates an 81-frame clip in 10.7\,s (7.6 frames/s), compared with 122\,s for the bidirectional model (\cref{tab:supp:speed}).

\subsection{Novel-view synthesis}
\label{sec:experiments:nvs}
\vspace{1mm}
\noindent\textbf{Protocol.}
We use 44 cases from PKU-DyMVHumans~\cite{zheng2024pkudymvhumans} where synchronized cameras provide ground-truth target views. To separate generation quality from monocular depth ambiguity, depth-based methods and our mesh condition use a subject distance triangulated from the camera ring, which is unavailable in monocular deployment. Framing and scoring windows are given in \cref{sec:supp:nvs_protocol}.

\begin{figure}[b]
\centering
\includegraphics[width=\linewidth]{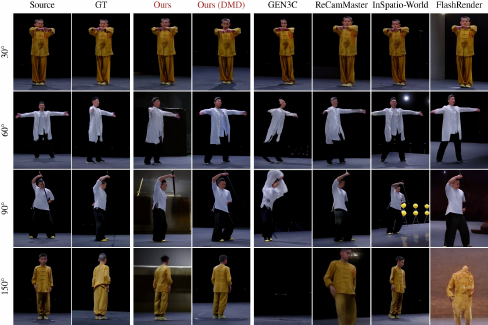}
\caption{\textbf{Novel-view synthesis on PKU-DyMVHumans.}
\ourmethod{} turns the person to match the ground-truth view at all angles, while the baselines increasingly fail to follow the viewpoint
as the angle grows, often keeping close to the source-facing pose. More methods are compared in \cref{fig:qual:scene_all}.}
\label{fig:qual:scene}
\end{figure}

\vspace{1mm}
\noindent\textbf{Metrics and results.}
Because PKU-DyMVHumans is captured in a dark studio with little background structure, all metrics are computed on the person region (details in \cref{sec:supp:nvs_protocol}).
\ourmethod{} (bidirectional) has the lowest orientation error and the best PSNR, SSIM, and LPIPS (\cref{tab:exp:nvs}, top), and among autoregressive methods, \ourmethod{}-DMD has lower orientation error than InSpatio-World.
\Cref{fig:qual:scene} shows the generated views against the calibrated target views.

\begin{table}[t]
\centering
\caption{\textbf{Novel-view synthesis on PKU-DyMVHumans.} Top: multi-view captures with calibrated target cameras; ReDirector and Geo-Align receive camera translations converted to their own metric scale ($\times k$). Bottom: $360^\circ$ rendering from a background-free video input. $^{\dagger}$Driven by GVHMR-estimated motion; orientation is scored with the same estimator. $^{\S}$Position and scale are given by the GT box, so these values (gray) are unranked. $^{\ddagger}$Fitted per source video.}
\label{tab:exp:nvs}
\footnotesize\setlength{\tabcolsep}{3pt}
\fitto{\columnwidth}{%
\begin{tabular}{lcccccc}\toprule
Method & Orient.$\downarrow$ & IoU$\uparrow$ & DINO$\uparrow$ & PSNR$\uparrow$ & SSIM$\uparrow$ & LPIPS$\downarrow$ \\
\midrule
\multicolumn{7}{l}{\textit{Multi-view captures}} \\
GEN3C~\cite{ren2025gen3c} & \underline{28.7} & 0.467 & 0.697 & 23.15 & 0.944 & 0.038 \\
ReCamMaster~\cite{bai2025recammaster} & 42.2 & \textbf{0.585} & 0.737 & \underline{24.89} & \underline{0.952} & \underline{0.030} \\
FlashRender~\cite{park2026flashrender} & 49.6 & 0.452 & 0.719 & 24.09 & 0.949 & 0.036 \\
ReDirector~\cite{park2025redirector} ($\times k$) & 52.6 & 0.309 & 0.723 & 22.01 & 0.944 & 0.045 \\
NeoVerse~\cite{yang2026neoverse} & 39.2 & 0.313 & 0.625 & 24.28 & 0.948 & 0.039 \\
Geo-Align~\cite{li2026geoalign} ($\times k$) & 44.7 & 0.301 & \underline{0.750} & 22.32 & 0.945 & 0.044 \\
TrajectoryCrafter~\cite{yu2025trajectorycrafter} & 61.5 & 0.317 & 0.553 & 23.46 & 0.948 & 0.039 \\
\ourmethod{} & \textbf{17.9} & \underline{0.578} & \textbf{0.794} & \textbf{25.44} & \textbf{0.953} & \textbf{0.029} \\
\cmidrule(lr){1-7}
InSpatio-World~\cite{inspatio2026world} & 47.4 & 0.517 & \textbf{0.803} & 23.99 & 0.948 & 0.033 \\
\ourmethod{}-DMD & \textbf{41.2} & \textbf{0.551} & 0.794 & \textbf{24.45} & \textbf{0.950} & \textbf{0.032} \\
\midrule
\multicolumn{7}{l}{\textit{$360^\circ$ rendering, video input}} \\
GEN3C~\cite{ren2025gen3c} & 51.5 & 0.502 & 0.598 & 18.05 & 0.925 & 0.094 \\
ReCamMaster~\cite{bai2025recammaster} & 56.2 & 0.292 & 0.599 & 18.94 & 0.932 & 0.079 \\
TrajectoryCrafter~\cite{yu2025trajectorycrafter} & 81.5 & 0.341 & 0.536 & 19.59 & \underline{0.934} & 0.074 \\
4DAnyone~\cite{jin4danyone}$^{\dagger}$ & 19.6 & \unr{0.686\tnmark{\S}} & \textbf{0.802} & \unr{22.78\tnmark{\S}} & \unr{0.941\tnmark{\S}} & \unr{0.048\tnmark{\S}} \\
LHM++~\cite{qiulhmplus}$^{\dagger}$ & \textbf{8.1} & \underline{0.594} & 0.765 & \underline{19.96} & \textbf{0.937} & \underline{0.073} \\
GauHuman~\cite{hugauhuman}$^{\ddagger}$ & 17.9 & 0.512 & 0.571 & 19.30 & 0.933 & 0.088 \\
\ourmethod{} & \underline{17.1} & \textbf{0.627} & \underline{0.780} & \textbf{20.20} & \underline{0.934} & \textbf{0.069} \\
\cmidrule(lr){1-7}
\ourmethod{}-DMD & 22.1 & 0.577 & 0.757 & 19.65 & 0.933 & 0.069 \\
\bottomrule\end{tabular}}
\end{table}

\subsection{Reconstruction via novel-view rendering}
\label{sec:experiments:reconstruction}
\looseness=-1
We evaluate 12 PKU subjects on white backgrounds along calibrated $360^\circ$ orbits: the input is an 81-frame monocular video,
targets are synchronized images from the camera ring, and depth-dependent inputs use the triangulated distance of \cref{sec:experiments:nvs}.
We report person IoU, DINO, white-background PSNR/SSIM/LPIPS with the GT person mask, and GVHMR orientation error (\cref{tab:exp:nvs}, bottom);
this measures appearance through rendered views, not mesh accuracy.
\ourmethod{} (bidirectional) reduces orientation error to $17.1^\circ$, compared with $51.5$--$81.5^\circ$ for the video-generation baselines, and has the best ranked IoU, PSNR, and LPIPS.
Explicit 3D methods remain competitive: LHM++ has lower orientation error and 4DAnyone higher pixel scores, under the conditions noted in \cref{tab:exp:nvs}.
Qualitative orbits, a single-image variant, and evaluation details are in \cref{fig:qual:video,sec:supp:single_image,sec:supp:reconstruction_checks}.

\subsection{Motion control}
\label{sec:experiments:motion}
\looseness=-1
We evaluate on 50 cases from RealisDance-Val~\cite{zhou2025realisdance}, UBC Fashion~\cite{zablotskaia2019dwnet}, and TikTok~\cite{jafarian2021tiktok}, using frame 0 as the reference image and frames 0--80 as driving motion and ground truth, and compare with Wan2.2-Animate-14B~\cite{cheng2025wananimate}, Wan2.2-Fun-5B-Control~\cite{alibabapai2025wanfun}, and One-to-All-1.3B~\cite{shi2026onetoall} (metrics and per-dataset results in \cref{sec:supp:motion50}).
\ourmethod{} is competitive with Wan2.2-Animate-14B, with the highest PSNR and SSIM.
Against the same-backbone Wan2.2-Fun-5B-Control, \ourmethod{} wins PSNR in 48 of 50 cases.

\begin{table}[t]
\centering
\caption{\textbf{Motion control.}
$^{\ddagger}$Wan2.2-Animate-14B additionally receives per-frame GT face crops from the driving video.
$^{\dagger}$One-to-All-1.3B was trained on UBC Fashion.}
\label{tab:exp:motion}
\footnotesize\setlength{\tabcolsep}{3pt}\renewcommand{\arraystretch}{0.88}\setlength{\aboverulesep}{0.8pt}\setlength{\belowrulesep}{0.8pt}\setlength{\cmidrulesep}{0pt}
\fitto{\columnwidth}{%
\begin{tabular}{lccccccc}\toprule
Method & PSNR$\uparrow$ & SSIM$\uparrow$ & LPIPS$\downarrow$ & ArcFace~\cite{deng2019arcface}$\uparrow$ & PCK@0.05$\uparrow$ & FID$\downarrow$ & FVD$\downarrow$ \\
\midrule
Wan2.2-Animate-14B~\cite{cheng2025wananimate}$^{\ddagger}$ & \underline{22.09} & \underline{0.757} & \textbf{0.148} & \textbf{0.467} & \textbf{0.927} & \textbf{24.6} & \textbf{170} \\
Wan2.2-Fun-5B-Control~\cite{alibabapai2025wanfun} & 19.49 & 0.701 & 0.218 & 0.250 & 0.867 & 42.8 & 291 \\
One-to-All-1.3B~\cite{shi2026onetoall}$^{\dagger}$ & 18.40 & 0.713 & 0.200 & \underline{0.365} & 0.856 & 37.7 & \underline{271} \\
\midrule
\ourmethod{} & \textbf{22.53} & \textbf{0.792} & \underline{0.155} & 0.280 & \underline{0.918} & \underline{34.2} & 272 \\
\bottomrule\end{tabular}}
\end{table}

\subsection{Joint camera and motion control}
\label{sec:experiments:joint_control}
\begin{figure}[t]
\centering
\includegraphics[width=\linewidth]{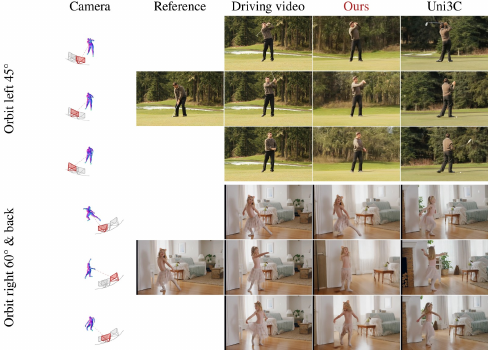}
\caption{\textbf{Joint camera and motion control.}
Uni3C follows the camera path more closely, but its person is often still seen from the wrong side, whereas \ourmethod{} shows the person from the commanded viewpoint. More cases are in \cref{fig:joint_control_all}.}
\label{fig:joint_control_visual}
\end{figure}

\begin{table}[!t]
\centering
\caption{\textbf{Joint camera and motion control.}}
\label{tab:exp:joint_motion}
\footnotesize\setlength{\tabcolsep}{3pt}\renewcommand{\arraystretch}{0.88}\setlength{\aboverulesep}{0.8pt}\setlength{\belowrulesep}{0.8pt}\setlength{\cmidrulesep}{0pt}
\fitto{\columnwidth}{%
\begin{tabular}{lccccccc}\toprule
& \multicolumn{3}{c}{Camera (VGGT)$\downarrow$} & \multicolumn{2}{c}{Motion (mm)$\downarrow$} & \multicolumn{2}{c}{Orient.\ ($^\circ$)$\downarrow$} \\
\cmidrule(lr){2-4}\cmidrule(lr){5-6}\cmidrule(lr){7-8}
Method & ATE & RPE & RRE ($^\circ$) & MPJPE & PA-MPJPE & Mean & Med. \\
\midrule
CamAnimate~\cite{wang2024humanvid} & 0.200 & 0.051 & 0.83 & \underline{63.30} & \underline{49.50} & \underline{21.49} & \underline{16.16} \\
RealisDance-DiT~\cite{zhou2025realisdance} & 0.200 & 0.036 & 0.74 & 75.91 & 59.62 & 28.98 & 28.84 \\
Uni3C~\cite{cao2025uni3c} & \textbf{0.047} & \textbf{0.013} & \textbf{0.25} & 89.27 & 66.61 & 35.94 & 35.37 \\
\midrule
\ourmethod{} & \underline{0.079} & \underline{0.022} & \underline{0.46} & \textbf{55.59} & \textbf{45.03} & \textbf{7.83} & \textbf{6.90} \\
\bottomrule\end{tabular}}
\end{table}

\vspace{1mm}
\noindent\textbf{Protocol.}
Following Uni3C~\cite{cao2025uni3c}, we compare with CamAnimate~\cite{wang2024humanvid}, RealisDance-DiT~\cite{zhou2025realisdance}, and Uni3C on 50 cases (10 driving videos $\times$ 5 trajectories), measuring camera accuracy with VGGT~\cite{wang2025vggt} and motion and orientation with GVHMR; details are in \cref{sec:supp:joint_checks,sec:supp:joint_quality}.

\vspace{1mm}
\noindent\textbf{Results.}
\ourmethod{} ranks first on human control (MPJPE and orientation following) and second on camera accuracy, behind Uni3C, whose person often fails to follow the commanded viewpoint (\cref{tab:exp:joint_motion,fig:joint_control_visual}); \ourmethod{}'s VBench quality is lower than the baselines' (\cref{tab:supp:joint_vbench}).
\subsection{Long-video generation}
\label{sec:experiments:long_video}
\looseness=-1
On 162-frame generation for 50 camera-control cases, \ourmethod{}-DMD has lower body-orientation error than InSpatio-World, and the full result is in \cref{sec:supp:long_video}.

\subsection{Ablation: conditioning representation}
\label{sec:experiments:ablation}
\looseness=-1
We replace the body-mesh renderings with per-pixel Plücker maps of the source and target cameras and train both variants for 10k steps under the same recipe (details in \cref{sec:supp:ablation}).
The Plücker variant shows the requested side of the body far less reliably, with lower joint success and higher orientation error (\cref{tab:supp:ablation}).

\section{Conclusion}
\label{sec:conclusion}
\looseness=-1
We presented \ourmethod{}, which anchors camera control to the human body: paired renderings of a body mesh along the source and target trajectories
specify the viewpoint change relative to the subject, and posed mesh sequences extend the same condition to motion control.
\ourmethod{} achieves the lowest body-orientation error and the highest joint success in camera control, follows the driving pose more closely
under joint camera and motion control, and supports a distilled causal variant that accepts camera commands during generation.

\vspace{1mm}
\looseness=-1
\noindent\textbf{Limitations.}
For motion control, the body condition does not encode facial expressions, and our training data contains no human--object interaction,
so \ourmethod{} cannot control expressions or make objects held in the reference image follow the hands.
Joint camera and motion control relies on a body mesh estimated from every driving frame, so it fails when the person is heavily occluded or only partly visible.

{
\small
\bibliographystyle{ieeenat_fullname}
\bibliography{main}
}

\clearpage
\appendix
\section*{Appendix}
\setcounter{figure}{0}\setcounter{table}{0}
\renewcommand{\thefigure}{S\arabic{figure}}
\renewcommand{\thetable}{S\arabic{table}}
\renewcommand{\theHfigure}{S\arabic{figure}}
\renewcommand{\theHtable}{S\arabic{table}}
\raggedbottom
\renewcommand{\dblfloatpagefraction}{0.5}\renewcommand{\floatpagefraction}{0.5}\renewcommand{\dbltopfraction}{0.9}\renewcommand{\topfraction}{0.9}\renewcommand{\textfraction}{0.1}\setcounter{dbltopnumber}{2}
This appendix provides implementation details and additional results. \Cref{sec:supp:data_construction} describes how the training pairs are constructed, and \cref{sec:supp:additional_details} gives the training schedule and the evaluation protocols for novel-view synthesis, reconstruction, and joint camera and motion control. \Cref{sec:camera_eval_details,sec:joint_sensitivity,sec:supp:motion_preservation} expand the camera-control evaluation: \cref{sec:camera_eval_details} describes the trajectories, camera estimation, and visual-quality metrics, \cref{sec:joint_sensitivity} defines the joint-success criterion and tests its sensitivity to the thresholds, and \cref{sec:supp:motion_preservation} measures how well the source motion is preserved. \Cref{sec:supp:single_image} reports novel-view synthesis from a single image, \cref{sec:supp:motion50} gives the motion-control evaluation details, and \cref{sec:supp:joint_quality} reports the visual quality of joint camera and motion control. \Cref{sec:supp:long_video} evaluates long-video generation and generation speed, and \cref{sec:supp:ablation} presents the conditioning ablation. \Cref{sec:supp:discussion} discusses in-the-wild use, including failure cases, anchor selection in multi-person scenes, and motion transferred from another video, and \cref{sec:supp:more_results} gives more visual results. Additional figures (\cref{fig:supp:mvhn_constructions,fig:qual:camera_all,fig:gen3c_pass_more,fig:joint_control_all,fig:qual:scene_all,fig:qual:video,fig:qual:image,fig:motion_formats,fig:supp:long_video,fig:supp:ablation_visual,fig:supp:ablation_nvs_visual,fig:supp:failures,fig:supp:anchor_choice,fig:supp:motion_transfer,fig:supp:more_scenes,fig:supp:more_orbits}) are placed in the corresponding sections.

\section{Training data construction}
\label{sec:supp:data_construction}
This section details the four data sources and the motion-control pairs summarized in \cref{sec:method:data} and \cref{fig:data_sources}. In every camera-control source, the condition uses a first-frame mesh held fixed over time; for MultiCamVideo (MCV)~\cite{bai2025recammaster} and DL3DV~\cite{ling2024dl3dv} one mesh is shared by all views, whereas for SynCam~\cite{bai2025syncammaster} and MVHumanNet~\cite{xiong2024mvhumannet} it is reconstructed independently in each view, so source and target meshes can differ slightly in shape and placement.

\vspace{1mm}
\noindent\textbf{MultiCamVideo.}
We sample distinct synchronized views with known camera intrinsics and trajectories. We track and segment the subject and fit human meshes to selected frames using the calibrated intrinsics. The static camera's first-frame mesh is the reference. Because the camera centers coincide initially, we use later moving-camera observations to resolve the reference mesh's depth. We place its pelvis along the reference viewing ray as $\mathbf{P}(\lambda)=\mathbf{C}_{\mathrm{ref}}+\lambda\mathbf{d}$, where $\mathbf{C}_{\mathrm{ref}}$ is the reference camera center and $\mathbf{d}$ is the unit ray through the estimated pelvis. To compare this static point with the moving views, we remove the subject's own motion. Per-frame pelvis estimates in each view are first brought to a common scale by rescaling them about the camera center by the ratio of the reference fit's rest-pose height to that frame's estimate, which changes only their depth, not their projection. The subject's displacement since frame 0, measured in the static camera, is then rotated into each moving camera and subtracted in 3D. The projections of these motion-compensated points serve as 2D targets, and a single $\lambda$ per scene minimizes a confidence-weighted soft-$L_1$ distance between them and the projection of $\mathbf{P}(\lambda)$ over all moving views and frames, excluding frame 0. Scaling the reference mesh about the camera center to the fitted depth preserves its initial projection. We then hold its pose and world position fixed and render it along each camera trajectory.

\vspace{1mm}
\noindent\textbf{SynCam.}
We sample synthetic multi-view video pairs and reconstruct a first-frame mesh independently in each view. Each mesh is rendered with its estimated camera, and the normal map is repeated over time with the same crop and resize as the RGB frames. We also assemble clips from multiple camera views of the same scene, choosing the shot counts independently for source and target and aligning cuts to latent-frame boundaries.

\vspace{1mm}
\noindent\textbf{MVHumanNet.}
MVHumanNet's studio backdrop offers few cues to camera motion. With the subject in isolation, a viewpoint change can resemble a body turn. To provide a visual cue, we extract subjects with the dataset masks and composite them onto textured indoor and outdoor scenes rendered from each calibrated viewpoint. We select angularly separated source and target views, reconstruct each view's normal map from its first frame, and apply crop-based camera motion and SynCam's multi-shot procedure.

\begin{figure}[t]
\centering
\includegraphics[width=\linewidth]{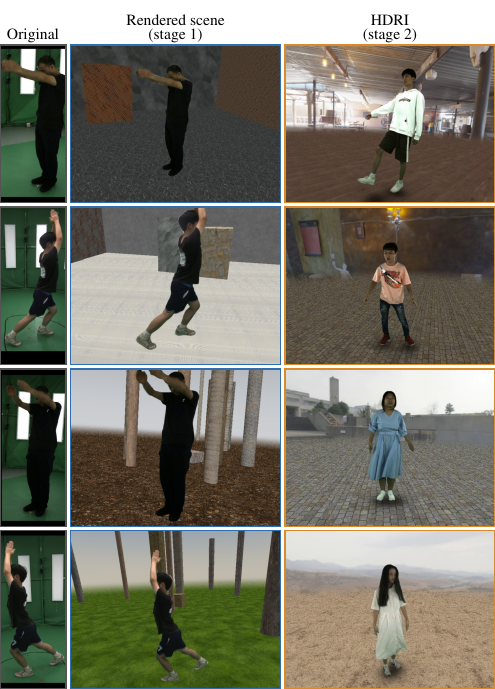}
\caption{\textbf{MVHumanNet training data.} Left: original studio frames. Middle: the same subjects composited onto rendered indoor and outdoor scenes (stage 1). Right: HDRI composites added in continued bidirectional training (stage 2); these show different subjects and are not paired with the other two columns of the same row.}
\label{fig:supp:mvhn_constructions}
\end{figure}

To broaden the range of backgrounds and give them a more photographic appearance, we add HDRI environments during continued bidirectional training. A textured ground plane provides near-field parallax. The target switches directly between calibrated views on the same elevation ring without interpolation, and each target frame uses the normal map for that camera. \Cref{fig:supp:mvhn_constructions} shows the original studio frames, the rendered-scene composites, and the HDRI composites; the HDRI examples are different subjects from the other two columns.

\vspace{1mm}
\noindent\textbf{DL3DV.}
DL3DV adds real scenes filmed along continuous camera orbits. Inserting a static person gives us examples in which camera motion alone changes the view of the body. We use SAM 3D Body~\cite{yang2026sam3dbody} to reconstruct an MHR mesh and texture it from multi-view MVHumanNet captures. Using the released calibration and a triangulated sparse point cloud, we avoid placing the mesh within scene geometry and favor positions that keep it visible along the trajectory. For each target frame, we composite the mesh onto the RGB image with a hard mask and render its normal map with the same camera. The composite omits cast shadows and occlusion by scene objects.

\vspace{1mm}
\noindent\textbf{Motion supervision.}
We derive motion-control pairs from MultiCamVideo (\cref{fig:data_sources}, bottom row) by repeating the static reference camera's first RGB frame and normal map as the source. Target meshes are reconstructed independently per frame in each view and rendered using the corresponding camera intrinsics and estimated mesh translation. Static and moving target cameras provide human-motion examples with and without camera motion.

Because both controls share one geometric condition, per-frame meshes can also be rendered under a moving target camera, so camera and motion control can be combined. We evaluate this setting in \cref{sec:experiments:joint_control}.

\section{Training and evaluation details}
\label{sec:supp:additional_details}
\vspace{1mm}
\noindent\textbf{Training schedule.}
All stages train rank-32 LoRA adapters on the attention and FFN projections of Wan2.2-Fun-5B-Control with AdamW (weight decay 0.01) on eight GPUs. The first bidirectional run uses MCV:SynCam:room-rendered MVHumanNet:DL3DV in a 7:1:1:1 mixture for 10,000 steps at batch size 32 and learning rate $10^{-4}$ with cosine decay. The continued run adds 20,000 steps on MCV camera-control, MVHumanNet-HDRI, and MCV motion-control data (2:2:3), so its final checkpoint has 30,000 total training steps. The causal variant starts from the 10,000-step checkpoint and is trained with Diffusion Forcing (DF)~\cite{chen2024diffusionforcing} for 10,000 steps on the initial mixture (batch size 32), then with DMD~\cite{dmd} for 5,000 steps (batch size 16) using the frozen 10,000-step teacher, four denoising steps per chunk, one generator update per five critic updates, and an EMA decay of 0.99; we report the EMA weights of the DMD student. All sampling uses guidance scale 1.0.

\vspace{1mm}
\noindent\textbf{Multi-view novel-view protocol.}
\label{sec:supp:nvs_protocol}
The PKU-DyMVHumans~\cite{zheng2024pkudymvhumans} test uses four nominal target angles ($30^\circ$, $60^\circ$, $90^\circ$, and $150^\circ$). Methods without depth input receive only the target camera trajectory, so the scale of its translation is ambiguous to them. By default we convert translations to meters with $k$, the meters per calibration unit implied by a 1.7\,m SAM 3D Body rest-pose height (the convention of the camera-control benchmark). Before the full run, two pilot cases per method were scored by person-mask IoU and center distance to the ground-truth view; a method uses unconverted calibration units only if both cases favored them. This held for ReCamMaster~\cite{bai2025recammaster} and FlashRender~\cite{park2026flashrender}; for ReDirector~\cite{park2025redirector} and Geo-Align~\cite{li2026geoalign} the two cases disagreed, so they keep the default (marked in \cref{tab:exp:nvs}). The pilot cases are from the test set; since the choice can only improve the baselines' placement, it does not favor \ourmethod{}. Framing uses a centered virtual pinhole camera with a subject height of about 400 pixels. For methods supporting smooth camera motion, including \ourmethod{}, the camera reaches the target view over the first 20 frames; all methods are scored on frames 21--80, or the corresponding 36 frames for 49-frame outputs. Methods limited to fixed-view outputs start at the target view directly.

\vspace{1mm}
\noindent\textbf{Triangulated distance.}
The shared distance in the calibrated novel-view tests is triangulated to the body center and applied as subject-surface depth for depth-dependent methods. This common convention removes one source of monocular scale ambiguity but does not make center and surface depth physically identical.

\begin{figure*}[t]
\centering
\includegraphics[width=\linewidth,height=.55\textheight,keepaspectratio]{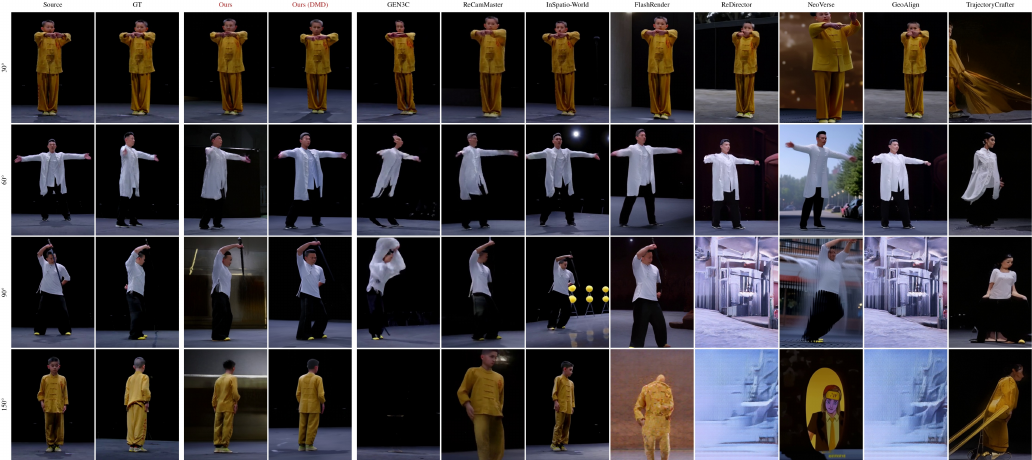}
\caption{\textbf{Novel-view synthesis on multi-view captures (PKU-DyMVHumans) with all compared methods.} This extends \cref{fig:qual:scene}. The calibrated target camera provides a reference for the subject's silhouette and placement relative to the background.}
\label{fig:qual:scene_all}
\end{figure*}

\vspace{1mm}
\noindent\textbf{Multi-view scoring.}
The ground-truth target image is mapped into each output's pixel frame through the two intrinsics, since the camera rotation and center coincide. Ground-truth person masks come from SAM~2~\cite{ravi2025sam2} prompted with a person box on the target view (the official PKU-DyMVHumans matte where SAM~2 fails). For PSNR, SSIM, and LPIPS, generated and ground-truth frames are both composited on white with the ground-truth person mask and compared over the full frame, so the shared white background raises these scores for every method; values are averaged over the scored frames of each case and then over the 44 cases. IoU compares this mask with a SAM~2 mask of the generated person, counting an empty mask as zero. DINO compares DINOv2~\cite{oquab2024dinov2} features of person-box crops (enlarged by 10\%, non-person pixels set to the ImageNet mean, resized to $224\times224$) at the same frame. Orientation error is the absolute difference in GVHMR~\cite{shen2024gvhmr} body yaw between generated and ground-truth frames, pooled over all frames valid in both rather than averaged per case.
\Cref{fig:qual:scene_all} extends \cref{fig:qual:scene} to all compared methods.

\begin{figure*}[t]
\centering
\includegraphics[width=\linewidth,height=.55\textheight,keepaspectratio]{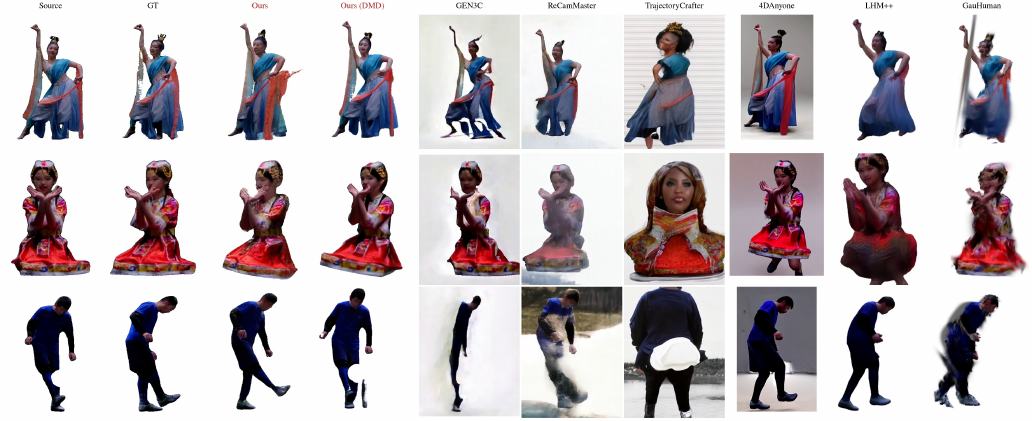}
\caption{\textbf{Human novel-view synthesis from a background-free video.} This accompanies the bottom part of \cref{tab:exp:nvs}. Views along the orbit show how the methods render the moving subject as different parts of the body come into view.}
\label{fig:qual:video}
\end{figure*}

\vspace{1mm}
\noindent\textbf{Reconstruction protocol.}
\label{sec:supp:reconstruction_checks}
Scoring excludes the repeated source views at the two orbit endpoints. Generated and ground-truth images are both composited on white with the ground-truth person mask, excluding hand-held objects, and body orientation is estimated by GVHMR independently on each.
4DAnyone~\cite{jin4danyone} and LHM++~\cite{qiulhmplus} are driven by GVHMR-estimated motion, the same estimator used for orientation scoring; other explicit 3D methods render their representations at the target cameras.
4DAnyone uses its nearest native view for each target keyframe (at most $3.8^\circ$ away), and its rendered person is placed on the ground-truth canvas using the ground-truth person box for position and scale, with uncovered pixels in the person region filled white (at most 13\% per frame).
Because this removes placement errors that the other methods must predict, its pixel scores and shape IoU are unranked, although it reaches PSNR 22.78\,dB, SSIM 0.941, and LPIPS 0.048 in \cref{tab:exp:nvs}, above both \ourmethod{} variants. These full-canvas scores include the white background, which every method matches. On ground-truth person pixels only (\cref{tab:supp:reconstruction_fg}), it still has the highest PSNR and lowest LPIPS, as expected from its ground-truth placement. Its orientation and DINO remain ranked, since neither depends on placement: orientation is estimated from each image, and DINO compares person-box crops resized to $224\times224$.
\Cref{fig:qual:video} shows the video-input orbits of \cref{sec:experiments:reconstruction}.

\begin{table}[!t]
\centering
\caption{\textbf{Foreground-only scores, video-input $360^\circ$ rendering.} Scores are computed only on pixels inside the ground-truth person mask. The 4DAnyone output is restored to the GT image canvas with GT-box-assisted placement; other methods retain their predicted placement. Red, orange, and yellow mark the first, second, and third best values. $^{\dagger}$GT-box-assisted placement.}
\label{tab:supp:reconstruction_fg}
\footnotesize\setlength{\tabcolsep}{4pt}
\begin{tabular}{lccc}\toprule
Method & PSNR (dB)$\uparrow$ & SSIM$\uparrow$ & LPIPS$\downarrow$ \\
\midrule
GEN3C~\cite{ren2025gen3c} & 7.13 & 0.218 & 0.627 \\
ReCamMaster~\cite{bai2025recammaster} & 7.72 & 0.184 & 0.737 \\
TrajectoryCrafter~\cite{yu2025trajectorycrafter} & 8.47 & 0.210 & 0.689 \\
4DAnyone~\cite{jin4danyone}$^{\dagger}$ & \cellcolor{red!22}11.56 & \cellcolor{yellow!18}0.237 & \cellcolor{red!22}0.495 \\
LHM++~\cite{qiulhmplus} & \cellcolor{yellow!18}8.74 & \cellcolor{red!22}0.263 & \cellcolor{yellow!18}0.568 \\
GauHuman~\cite{hugauhuman} & 8.08 & 0.227 & 0.634 \\
\midrule
\ourmethod{} & \cellcolor{orange!20}8.98 & \cellcolor{orange!20}0.240 & \cellcolor{orange!20}0.546 \\
\ourmethod{}-DMD & 8.43 & 0.207 & 0.570 \\
\bottomrule\end{tabular}
\end{table}

\vspace{1mm}
\noindent\textbf{Joint-control diagnostics.}
\label{sec:supp:joint_checks}
We follow the baseline and conditioning choices of Uni3C's unified-control evaluation~\cite{cao2025uni3c}: CamAnimate~\cite{wang2024humanvid} uses the driving video's 2D poses (with its released closed-loop pipeline), RealisDance-DiT~\cite{zhou2025realisdance} receives SMPL-X rendered under the target camera, and we also run Uni3C.
Ten driving videos from RealisDance-Val~\cite{zhou2025realisdance} and X-Dance are each paired with five trajectories: a $45^\circ$ horizontal orbit, a $60^\circ$ horizontal orbit with return, a $60^\circ$ vertical orbit with return, zoom-out with rightward pan, and zoom-out with upward pedestal motion, giving 50 cases evaluated in a $768\times480$ landscape or $480\times768$ portrait window.
Image-space pan and pedestal commands use each method's focal length, so the resulting physical trajectories can differ.
Uni3C's unified-control test set is not public and its trajectory normalization differs from ours, so its published numbers are not directly comparable.

Camera accuracy uses VGGT~\cite{wang2025vggt} ATE, RPE, and RRE over all 81 frames.
Orientation following uses command-adjusted GVHMR body yaw.
Because body yaw alone cannot distinguish a camera orbit from the subject turning in place, we also estimate the camera from the background only (ViPE~\cite{huang2025vipe} with person masking).
On the 20 horizontal orbit cases, background rotation reach (1 = full commanded rotation) is 0.947 for Uni3C, 0.888 for \ourmethod{}, 0.712 for CamAnimate, and 0.278 for RealisDance-DiT, and command-relative rotation error is $3.11^\circ$, $7.74^\circ$, $10.56^\circ$, and $14.53^\circ$, respectively (\cref{tab:supp:joint_background}).
RealisDance-DiT thus rotates the background little despite its low full-frame RRE ($0.74^\circ$), and the background estimates preserve the ordering of Uni3C and \ourmethod{} in the main table.

\begin{table}[!t]
\centering
\caption{\textbf{Background-camera diagnostics for joint control.} Lower errors are better; reach ratios are best at 1.}
\label{tab:supp:joint_background}
\footnotesize\setlength{\tabcolsep}{3pt}
\fitto{\columnwidth}{%
\begin{tabular}{llcccc}\toprule
Est. & Metric & \ourmethod{} & Uni3C~\cite{cao2025uni3c} & CamAnimate~\cite{wang2024humanvid} & RealisDance-DiT~\cite{zhou2025realisdance} \\
\midrule
ViPE~\cite{huang2025vipe} & ATE$\downarrow$ & \underline{0.091} & \textbf{0.036} & 0.207 & 0.189 \\
 & RPE$\downarrow$ & \underline{0.012} & \textbf{0.005} & 0.034 & 0.024 \\
 & RRE ($^\circ$)$\downarrow$ & \underline{0.43} & \textbf{0.20} & 0.97 & 0.73 \\
 & Cmd.-rel.\ rot.\ ($^\circ$)$\downarrow$ & \underline{7.74} & \textbf{3.11} & 10.56 & 14.53 \\
 & Horiz.\ reach ($\to1$) & \underline{0.888} & \textbf{0.947} & 0.712 & 0.278 \\
 & Vert.\ reach ($\to1$) & \underline{0.623} & \textbf{0.779} & 0.466 & 0.082 \\
\bottomrule\end{tabular}}
\end{table}

\section{Camera-control evaluation details}
\label{sec:camera_eval_details}
\vspace{1mm}
\noindent\textbf{Trajectories and inputs.}
The 11 Standard trajectories are Pan left/right $20^\circ$, Tilt up/down $10^\circ$, Zoom in/out (to 0.5 and 1.5 times the initial distance), Orbit up $60^\circ$, and Orbit left/right $45^\circ$ and $90^\circ$. The 7 Extended trajectories are Orbit up $60^\circ$ \& back, Orbit left $45^\circ\to$ right $45^\circ$, Orbit $60^\circ$ \& back, Orbit $90^\circ$ \& back, Orbit up $60^\circ$ + $90^\circ$, Roll $180^\circ$, and a large lateral Truck. Each method is evaluated on the same manifest and receives the same crop and per-frame trajectory through method-specific adapters. All methods receive the same trajectories, defined in the ViPE-based source frame under the same body-height scale convention. Methods with their own depth estimator keep it, and their commanded translations are rescaled by the median ratio of their frame-0 depth to the ViPE depth outside the person mask; only GEN3C~\cite{ren2025gen3c} consumes the ViPE depth and source poses directly.

\vspace{1mm}
\noindent\textbf{Visual-quality metrics.}
MEt3R~\cite{asim2025met3r} compares generated and source frames at matching times, sampled every fourth frame. The five VBench~\cite{huang2023vbench} dimensions are subject consistency, background consistency, motion smoothness, aesthetic quality, and imaging quality; Overall is their equal-weight mean.

\begin{table}[!t]
\centering
\caption{\textbf{Visual quality on in-the-wild human videos.} Overall: mean of the five VBench dimensions. $^{\ddagger}$TrajectoryCrafter outputs 49 frames, so its FVD (gray) is unranked.}
\label{tab:supp:quality}
\footnotesize\setlength{\tabcolsep}{2pt}
\fitto{\columnwidth}{%
\begin{tabular}{lccccccccc}\toprule
& & \multicolumn{6}{c}{VBench$\uparrow$} & & \\
\cmidrule(lr){3-8}
Method & MEt3R$\downarrow$ & Overall & Subj. & Bg. & Smooth. & Aesth. & Img. & FID$\downarrow$ & FVD$\downarrow$ \\
\midrule
ReCamMaster & 0.267 & 0.815 & \cellcolor{orange!20}0.911 & 0.908 & \cellcolor{orange!20}0.989 & \cellcolor{yellow!18}0.584 & 0.684 & 58.65 & \cellcolor{red!22}719 \\
TrajectoryCrafter & 0.348 & 0.786 & 0.871 & 0.896 & 0.973 & 0.541 & 0.651 & 74.41 & \unr{1114\tnmark{\ddagger}} \\
ReDirector & 0.281 & 0.814 & 0.897 & 0.919 & 0.985 & 0.575 & 0.694 & 59.35 & 836 \\
GEN3C & \cellcolor{orange!20}0.250 & 0.812 & 0.904 & \cellcolor{yellow!18}0.925 & \cellcolor{red!22}0.990 & 0.574 & 0.668 & 58.80 & 794 \\
DepthDirector & 0.313 & 0.782 & 0.860 & 0.903 & \cellcolor{yellow!18}0.988 & 0.542 & 0.618 & 75.57 & 981 \\
Geo-Align & 0.267 & \cellcolor{yellow!18}0.817 & 0.904 & 0.922 & 0.986 & 0.577 & \cellcolor{yellow!18}0.697 & \cellcolor{yellow!18}56.62 & 807 \\
FlashRender & \cellcolor{yellow!18}0.258 & \cellcolor{orange!20}0.824 & \cellcolor{yellow!18}0.908 & \cellcolor{orange!20}0.926 & 0.985 & \cellcolor{orange!20}0.589 & \cellcolor{red!22}0.709 & \cellcolor{orange!20}53.60 & \cellcolor{yellow!18}787 \\
NeoVerse & 0.298 & 0.811 & 0.893 & 0.918 & 0.986 & 0.553 & \cellcolor{orange!20}0.705 & 68.92 & 940 \\
\ourmethod{} & 0.262 & 0.816 & 0.901 & 0.923 & \cellcolor{yellow!18}0.988 & 0.576 & 0.694 & 58.35 & 831 \\
\midrule
InSpatio-World & 0.285 & 0.806 & 0.895 & 0.898 & 0.984 & 0.570 & 0.684 & 67.00 & 893 \\
\ourmethod{}-DMD & \cellcolor{red!22}0.242 & \cellcolor{red!22}0.828 & \cellcolor{red!22}0.917 & \cellcolor{red!22}0.931 & \cellcolor{yellow!18}0.988 & \cellcolor{red!22}0.597 & \cellcolor{red!22}0.709 & \cellcolor{red!22}52.40 & \cellcolor{orange!20}768 \\
\bottomrule\end{tabular}}
\end{table}

\vspace{1mm}
\noindent\textbf{Trajectory construction.}
The world frame is the initial source camera. The subject pivot uses the initial mask center and median ViPE~\cite{huang2025vipe} depth, and mesh scale assumes a 1.7\,m rest-pose body height (source depths and translations share this scale; see below). This height fixes a common scale that makes camera translations approximately metric for methods that consume them without depth input; all evaluation trajectories are defined by angles, distance ratios, or image-space constraints, rather than translations in meters. Zoom moves to 0.5 or 1.5 times the initial camera--subject distance with a fixed focal length, the vertical orbit turns by $60^\circ$, and lateral translation places the outermost projected mesh point 16 pixels from the image boundary.

\vspace{1mm}
\noindent\textbf{Scale invariance and verification.}
Changing the assumed height from $h$ to $h^{\prime}$ rescales source depth, source camera translations, the subject pivot, and commanded camera translations by the same factor $k=h^{\prime}/h$. Perspective projection is unchanged under this joint rescaling, so expected shoulder positions, projected subject sizes and visibility checks are invariant. Human-yaw and camera-rotation errors use rotations only; camera-translation error uses Sim(3) alignment and normalization by trajectory span. We verified the construction with height assumptions of 1.5, 1.7, and 1.9\,m. Re-exported trajectories matched the consistently rescaled translations within $10^{-9}$; expected shoulder positions shifted by a median of approximately $10^{-8}H$. No frame-level visibility states or case-level joint-success decisions changed, metric differences were at most $5\times10^{-9}$, and method rankings were unchanged.

\vspace{1mm}
\noindent\textbf{Orientation tracking.}
For in-the-wild camera-control evaluation, GVHMR provides orientation whenever it tracks a person. For each case, mean orientation error averages frames tracked in both source and output; the reported score then averages these case means. SAM 3D Body and 4D-Humans~\cite{goel2023humans4d} remain auxiliary estimators with their own valid supports (\cref{tab:supp:orient_estimators}). \Cref{tab:supp:orient_significance} gives paired source-level bootstrap intervals and sign tests for the orientation error against each method. F/S/B accuracy classifies the GVHMR body yaw $\psi$ relative to the camera as front ($|\psi|\le45^\circ$), side ($45^\circ<|\psi|<135^\circ$), or back ($|\psi|\ge135^\circ$), and counts a frame as correct when the output's class matches that of the expected yaw (source yaw plus commanded rotation). It uses the same frames and orbit cases as the orientation error and is averaged over cases.

\begin{table}[!t]
\centering
\caption{\textbf{Body-orientation error with three estimators.} Mean error ($^\circ$, lower is better) on orbit cases of the camera-control benchmark. Rows below the rule are autoregressive. The three estimators give consistent rankings.}
\label{tab:supp:orient_estimators}
\footnotesize\setlength{\tabcolsep}{4pt}
\begin{tabular}{lccc}\toprule
Method & GVHMR & SAM 3D Body & 4D-Humans \\
\midrule
ReCamMaster~\cite{bai2025recammaster} & 19.71 & 18.70 & 18.44 \\
TrajectoryCrafter~\cite{yu2025trajectorycrafter} & 39.85 & 39.92 & 39.02 \\
ReDirector~\cite{park2025redirector} & 19.82 & 17.45 & 19.21 \\
GEN3C~\cite{ren2025gen3c} & 20.52 & 20.39 & 21.34 \\
DepthDirector~\cite{chen2026depthdirector} & 28.84 & 23.15 & 21.51 \\
Geo-Align~\cite{li2026geoalign} & 21.33 & 17.12 & 20.84 \\
FlashRender~\cite{park2026flashrender} & \underline{16.99} & \underline{15.80} & \underline{18.00} \\
NeoVerse~\cite{yang2026neoverse} & 35.27 & 32.77 & 31.76 \\
\ourmethod{} & \textbf{12.61} & \textbf{13.09} & \textbf{13.81} \\
\midrule
InSpatio-World~\cite{inspatio2026world} & 23.50 & 21.80 & 23.05 \\
\ourmethod{}-DMD & \textbf{20.25} & \textbf{20.12} & \textbf{20.04} \\
\bottomrule\end{tabular}
\end{table}

\begin{table}[t]
\centering
\caption{\textbf{Significance of the orientation results.} \ourmethod{} minus each method in GVHMR body-orientation error ($^\circ$) on the orbit cases of \cref{tab:exp:camera}; negative favors \ourmethod{}. Brackets: 95\% paired bootstrap interval over source videos (10{,}000 replicates). $p$: two-sided exact sign test over source videos; all comparisons remain significant after Holm correction (largest Holm-adjusted $p=0.024$). $^{\ast}$Autoregressive.}
\label{tab:supp:orient_significance}
\footnotesize\setlength{\tabcolsep}{3pt}
\fitto{\columnwidth}{%
\begin{tabular}{lccc}\toprule
\ourmethod{} vs.\ & Ours / other & $\Delta$ [95\% CI] & $p$ \\
\midrule
FlashRender~\cite{park2026flashrender} & 12.61 / 16.99 & $-4.38$ [$-7.38$, $-1.14$] & 0.012 \\
GEN3C~\cite{ren2025gen3c} & 12.61 / 20.52 & $-7.91$ [$-11.87$, $-4.32$] & 0.012 \\
ReCamMaster~\cite{bai2025recammaster} & 12.61 / 19.71 & $-7.10$ [$-10.13$, $-3.52$] & $<$0.001 \\
TrajectoryCrafter~\cite{yu2025trajectorycrafter} & 12.61 / 39.85 & $-27.24$ [$-34.64$, $-20.07$] & $<$0.001 \\
ReDirector~\cite{park2025redirector} & 12.61 / 19.82 & $-7.21$ [$-10.53$, $-4.12$] & $<$0.001 \\
DepthDirector~\cite{chen2026depthdirector} & 12.61 / 28.84 & $-16.23$ [$-23.13$, $-10.37$] & 0.001 \\
Geo-Align~\cite{li2026geoalign} & 12.61 / 21.33 & $-8.72$ [$-12.36$, $-5.22$] & $<$0.001 \\
NeoVerse~\cite{yang2026neoverse} & 12.61 / 35.27 & $-22.66$ [$-34.50$, $-13.19$] & $<$0.001 \\
\midrule
InSpatio-World~\cite{inspatio2026world}$^{\ast}$ & 12.61 / 23.50 & $-10.90$ [$-16.75$, $-5.89$] & 0.001 \\
\ourmethod{}-DMD$^{\ast}$ & 12.61 / 20.25 & $-7.64$ [$-10.41$, $-4.64$] & $<$0.001 \\
\bottomrule\end{tabular}}
\end{table}

\begin{figure*}[!p]
\centering
\makebox[\linewidth][c]{\includegraphics[width=.94\paperwidth,height=.88\textheight,keepaspectratio]{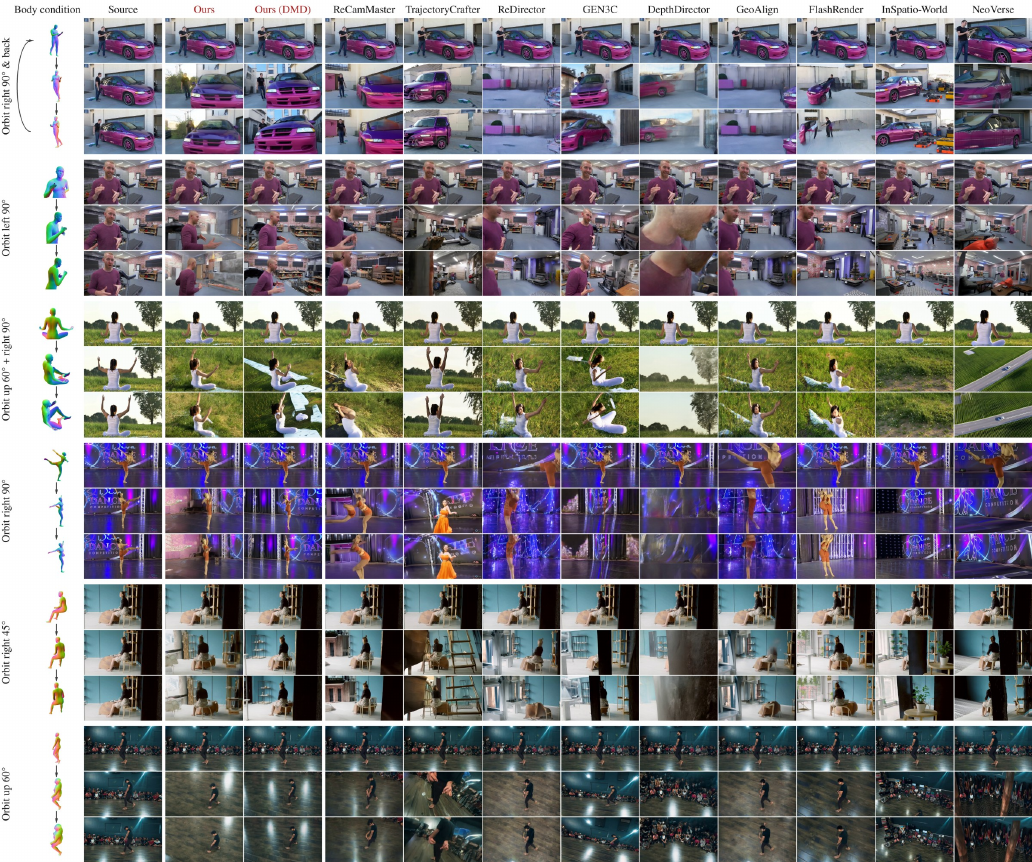}}
\caption{\textbf{Camera control on human videos with all compared methods.} This extends \cref{fig:qual:camera}. Changes in the background indicate camera motion, while the face and torso show whether the person is seen from the requested side.}
\label{fig:qual:camera_all}
\end{figure*}

\begin{figure*}[p]
\centering
\includegraphics[width=\linewidth,height=.78\textheight,keepaspectratio]{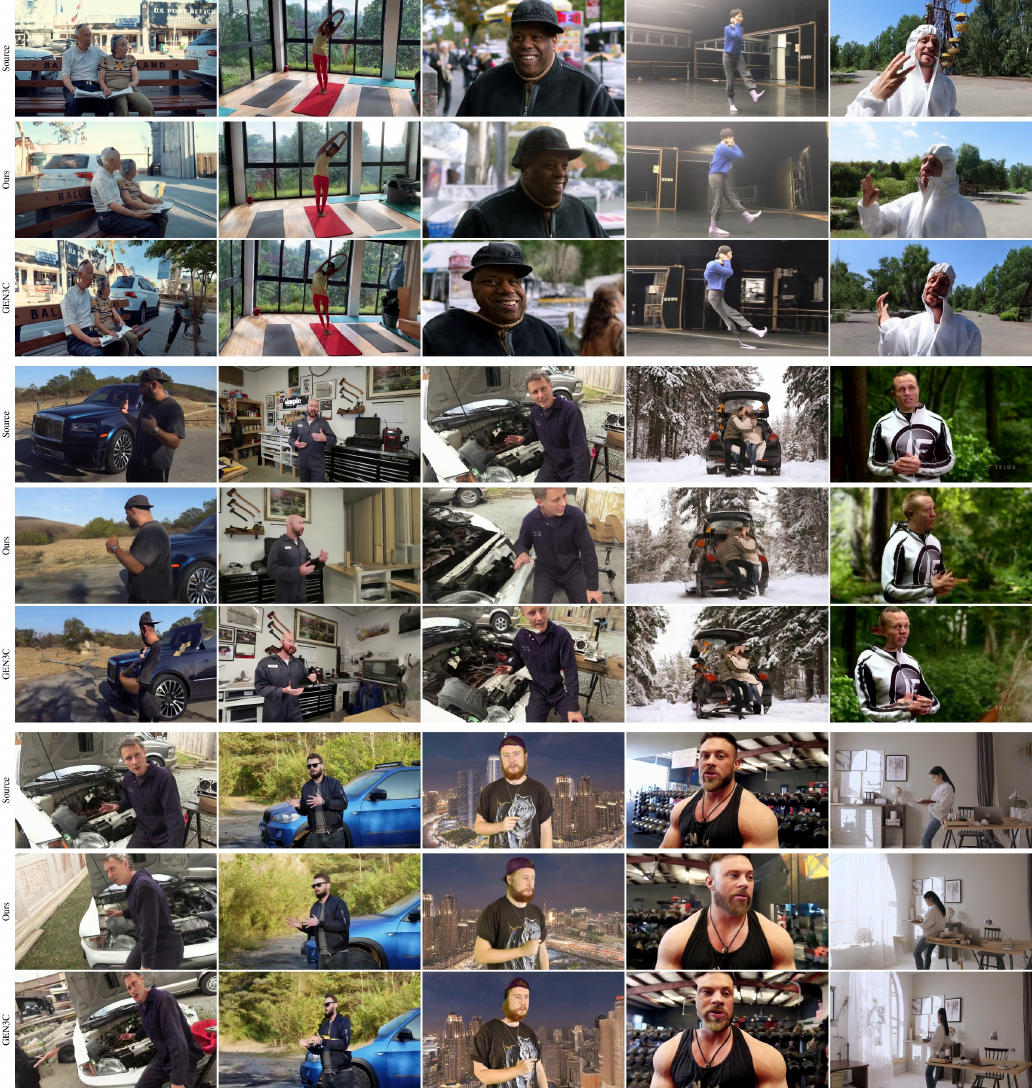}
\caption{\textbf{Body shape in cases where both \ourmethod{} and GEN3C pass the joint camera-control evaluation.} The paired views show differences in body proportions that the pass rate alone cannot capture.}
\label{fig:gen3c_pass_more}
\end{figure*}

\vspace{1mm}
\noindent\textbf{Position and missing rate.}
Orientation and position errors average valid frames within each case. Missing rate pools eligible frames across cases. Source frames must have both shoulders and hips detected to estimate depth; the same source-side exclusions apply to every method. On the output side, only the two shoulders are required, each at confidence at least 0.5. Expected visibility is determined by whether the projected shoulder midpoint lies inside the image. Position error is its displacement from the detected output shoulder midpoint divided by image height $H$. Missing rate counts eligible frames with an in-image expected shoulder midpoint but no valid output shoulder pair. \Cref{fig:qual:camera_all} extends \cref{fig:qual:camera} to all compared methods on the same camera-control cases, and \cref{fig:gen3c_pass_more} shows 15 cases in which both \ourmethod{} and GEN3C pass the joint evaluation.

\section{Joint-success evaluation}
\label{sec:joint_sensitivity}
We evaluate the combined 580-case set with tracking-based GVHMR orientation: a tracked person provides a yaw reading without a shoulder-and-hip visibility requirement. \Cref{tab:exp:joint} reports joint success separately for each split.

\vspace{1mm}
\noindent\textbf{Threshold definitions.}
Let $C,O,P,B$ denote camera, orientation, person, and blur failures; joint success requires $\neg(C\lor O\lor P\lor B)$ over all applicable checks, with baseline parameters $(k_c,k_o,p,\delta)=(0.5,0.5,0.10,0.2)$. Camera failure uses ViPE~\cite{huang2025vipe}: on rotational paths it is $e_{\mathrm{ViPE}}>\max(3^\circ,k_c e^0_{\mathrm{ViPE}})$, where $e^0_{\mathrm{ViPE}}$ is computed with the same scoring function from the commanded path under no camera motion, and return paths also apply this check to the turning-point segment (five frames on each side). On translational paths, camera failure is a Sim(3)-aligned translation error, normalized by trajectory span, above 0.15, or a wrong direction (box-height change for zoom, position-change direction for lateral motion). Body-orientation failure on applicable orbit paths, including return segments, is $e_o\geq\max(10^\circ,k_o e_o^0)$, where $e_o^0$ is the error of a non-turning human under the commanded camera. Person failure is a shoulder-detection fraction below 0.5, an orientation-tracked fraction below 0.5, or a case-mean normalized position error $e_p/H\geq p$; output visibility requires both shoulders at confidence at least 0.5, and expected visibility uses the projected shoulder midpoint. Blur failure is $q_{\mathrm{src}}-q_{\mathrm{out}}>\delta$ in VBench imaging quality. In every setting, the following stay fixed: the $3^\circ$ and $10^\circ$ floors, the translation-error limit of 0.15, the direction checks (box-height change for zoom and position-change direction for lateral motion), and the source-side shoulder-and-hip checks for depth, which are shared across methods. Numerical thresholds were set from GT measurement error and static references before joint success was computed; earlier per-method averages had already been inspected, and later tracking and visibility revisions kept these thresholds.

\vspace{1mm}
\noindent\textbf{Threshold sensitivity.}
We recompute joint success from the same per-case scores under 11 threshold settings, using the final bidirectional checkpoint and the DMD variant of \ourmethod{} throughout (\cref{tab:sens:Standard_Extended}). Single-factor settings vary one parameter at a time; all-tight uses $(0.3,0.3,0.05,0.1)$ and all-loose uses $(0.7,0.7,0.15,0.3)$ for $(k_c,k_o,p,\delta)$. \ourmethod{} has the highest joint success in 10 of the 11 settings; under the tightest camera threshold ($k_c=0.3$), GEN3C is higher (74.31\% vs.\ 68.45\%).

\begin{table*}[!t]
\centering
\caption{\textbf{Tracking-based GVHMR joint-success sensitivity on the combined 580-case set (\%$\uparrow$).} Rows below the rule are autoregressive.}
\label{tab:sens:Standard_Extended}
\small\setlength{\tabcolsep}{4pt}
\fitto{\textwidth}{%
\begin{tabular}{lccccccccccc}\toprule
& & \multicolumn{2}{c}{Camera thr.} & \multicolumn{2}{c}{Orientation thr.} & \multicolumn{2}{c}{Position thr.} & \multicolumn{2}{c}{Blur thr.} & \multicolumn{2}{c}{All thresholds} \\
\cmidrule(lr){3-4}\cmidrule(lr){5-6}\cmidrule(lr){7-8}\cmidrule(lr){9-10}\cmidrule(lr){11-12}
Method & Base & $0.3\times$ & $0.7\times$ & $0.3\times$ & $0.7\times$ & $0.05H$ & $0.15H$ & 0.1 & 0.3 & Tight & Loose \\
\midrule
ReCamMaster~\cite{bai2025recammaster} & 42.24 & 35.69 & 42.93 & 33.97 & 48.28 & 23.62 & 52.93 & 39.83 & 42.41 & 15.69 & 61.21 \\
TrajectoryCrafter~\cite{yu2025trajectorycrafter} & 40.34 & 39.66 & 40.69 & 39.48 & 40.69 & 35.34 & 45.00 & 37.59 & 40.34 & 31.72 & 46.03 \\
ReDirector~\cite{park2025redirector} & 55.52 & 55.17 & 55.69 & 52.41 & 56.21 & 34.14 & 64.14 & 55.00 & 55.52 & 32.59 & 65.34 \\
GEN3C~\cite{ren2025gen3c} & \underline{75.52} & \textbf{74.31} & \underline{75.86} & \underline{68.79} & \underline{77.59} & \underline{64.31} & \underline{78.97} & \underline{68.10} & \underline{75.69} & \underline{53.45} & \underline{82.24} \\
DepthDirector~\cite{chen2026depthdirector} & 48.10 & 46.72 & 48.97 & 46.03 & 48.62 & 40.52 & 52.07 & 40.52 & 48.28 & 32.93 & 54.31 \\
Geo-Align~\cite{li2026geoalign} & 55.00 & 54.83 & 55.17 & 51.55 & 56.03 & 36.55 & 63.45 & 54.14 & 55.00 & 34.48 & 65.17 \\
FlashRender~\cite{park2026flashrender} & 62.07 & 61.55 & 62.07 & 55.69 & 63.62 & 39.14 & 71.55 & 61.21 & 62.07 & 34.66 & 73.28 \\
NeoVerse~\cite{yang2026neoverse} & 52.24 & 52.07 & 52.24 & 50.17 & 55.69 & 40.52 & 58.45 & 52.07 & 52.24 & 38.97 & 62.41 \\
\ourmethod{} & \textbf{78.62} & \underline{68.45} & \textbf{80.52} & \textbf{75.17} & \textbf{78.97} & \textbf{68.79} & \textbf{82.24} & \textbf{75.17} & \textbf{78.97} & \textbf{55.86} & \textbf{84.83} \\
\midrule
InSpatio-World~\cite{inspatio2026world} & 58.10 & \textbf{57.07} & 58.28 & \textbf{56.03} & 59.48 & 47.93 & 63.28 & 53.79 & 58.45 & \textbf{43.45} & 65.69 \\
\ourmethod{}-DMD & \textbf{65.52} & 52.76 & \textbf{67.41} & 53.45 & \textbf{70.17} & \textbf{56.55} & \textbf{69.14} & \textbf{65.34} & \textbf{65.69} & 41.55 & \textbf{77.93} \\
\bottomrule\end{tabular}}
\end{table*}

\begin{table}[!t]
\centering
\caption{\textbf{Joint success by split.} Joint success uses tracking-based GVHMR; the All column is reported in \cref{tab:exp:camera}.}
\label{tab:exp:joint}
\footnotesize\setlength{\tabcolsep}{4pt}
\begin{tabular}{lccc}\toprule
& \multicolumn{3}{c}{Joint success (\%)$\uparrow$} \\
\cmidrule(lr){2-4}
Method & Std. & Ext. & All \\
\midrule
ReCamMaster~\cite{bai2025recammaster} & 44.32 & 35.71 & 42.24 \\
TrajectoryCrafter~\cite{yu2025trajectorycrafter} & 47.95 & 16.43 & 40.34 \\
ReDirector~\cite{park2025redirector} & 61.14 & 37.86 & 55.52 \\
GEN3C~\cite{ren2025gen3c} & \underline{76.82} & \underline{71.43} & \underline{75.52} \\
DepthDirector~\cite{chen2026depthdirector} & 54.09 & 29.29 & 48.10 \\
Geo-Align~\cite{li2026geoalign} & 60.68 & 37.14 & 55.00 \\
FlashRender~\cite{park2026flashrender} & 65.45 & 51.43 & 62.07 \\
NeoVerse~\cite{yang2026neoverse} & 57.95 & 34.29 & 52.24 \\
\ourmethod{} & \textbf{80.23} & \textbf{73.57} & \textbf{78.62} \\
\midrule
InSpatio-World~\cite{inspatio2026world} & 62.27 & 45.00 & 58.10 \\
\ourmethod{}-DMD & \textbf{69.55} & \textbf{52.86} & \textbf{65.52} \\
\bottomrule\end{tabular}
\end{table}

\section{Motion preservation under camera control}
\label{sec:supp:motion_preservation}
Camera control should change the viewpoint without changing what the person does. On the camera-control benchmark, we run GVHMR on the source and generated videos and compare the body pose frame by frame, with global orientation and translation removed: MPJPE over the 21 non-pelvis joints, PA-MPJPE after similarity alignment, and ROT, the mean geodesic angle between local rotations of eight limb joints. Frames need both shoulders and hips at confidence $\geq0.5$, and cases with fewer than 20 valid frames are not computable. We report orbit cases in \cref{tab:supp:motion_main} and non-orbit and tilt cases in \cref{tab:supp:motion_other}.

A low pose error alone does not show good camera control: if the view is not rotated, the estimator sees the body from the source viewpoint and the error drops. On orbit cases we therefore also check whether the body turns with the camera, using the orientation criterion of \cref{sec:joint_sensitivity}: the mean GVHMR facing-direction error, relative to the source direction plus the commanded camera rotation, fails when $e_o\geq\max(10^\circ, k_o e_o^0)$ with $k_o=0.5$. Since a body that does not turn has error $|\varphi_t|$ at each frame $t$, where $\varphi_t$ is the commanded camera rotation relative to the source, $e_o^0$ equals the mean commanded rotation over the scored frames. A body that does not turn, turns the opposite way, or turns much more than commanded fails this check. For return trajectories, the frames around the turnaround are checked with the same rule. As in the joint success rate, this check uses every frame in which the tracker finds the person, without the shoulder and hip confidence requirement that the pose metrics above use.

GVHMR itself returns different poses when the same motion is seen from a different viewpoint. We measure this viewpoint noise on synchronized PKU-DyMVHumans captures (4 subjects $\times$ 4 source cameras, 16 view pairs per separation), comparing two views of the same instant (\cref{tab:supp:motion_noise}). For each orbit case, the excess is its MPJPE minus the median noise at the case's angle (linearly interpolated), computed only on cases that pass the orientation check.

\begin{table}[t]
\centering
\caption{\textbf{Viewpoint noise of the pose estimator.} Pose difference between GVHMR estimates from two synchronized PKU-DyMVHumans views of the same instant (16 view pairs, about 1{,}270 valid frames per separation). MPJPE and PA-MPJPE in mm, ROT in degrees. The $0^\circ$ row reruns the same video.}
\label{tab:supp:motion_noise}
\footnotesize\setlength{\tabcolsep}{4pt}
\begin{tabular}{lcccc}\toprule
Separation & MPJPE mean & MPJPE median & PA-MPJPE & ROT \\
\midrule
$0^\circ$ & 0.00 & 0.00 & 0.00 & 0.01 \\
$29.9^\circ$ & 43.70 & 37.63 & 32.35 & 8.97 \\
$61.1^\circ$ & 64.40 & 53.01 & 50.18 & 12.97 \\
$90.5^\circ$ & 71.84 & 60.72 & 56.84 & 14.74 \\
$148.9^\circ$ & 58.02 & 47.96 & 45.27 & 12.92 \\
\bottomrule\end{tabular}
\end{table}

\begin{table}[t]
\centering
\caption{\textbf{Motion preservation on orbit trajectories.} MPJPE and PA-MPJPE (mm) and ROT ($^\circ$) averaged over cases; Orient.: body-orientation error ($^\circ$); Wrong: share of cases failing the orientation check of \cref{sec:joint_sensitivity} ($e_o\geq\max(10^\circ, 0.5\,e_o^0)$); Excess: MPJPE minus median viewpoint noise on cases passing the orientation check.}
\label{tab:supp:motion_main}
\footnotesize\setlength{\tabcolsep}{3pt}
\fitto{\columnwidth}{%
\begin{tabular}{lcccccc}\toprule
Method & MPJPE$\downarrow$ & PA$\downarrow$ & ROT$\downarrow$ & Orient.$\downarrow$ & Wrong (\%)$\downarrow$ & Excess$\downarrow$ \\
\midrule
ReCamMaster~\cite{bai2025recammaster} & \textbf{55.57} & \underline{45.12} & \textbf{12.33} & 19.71 & 32.49 & \textbf{12.91} \\
TrajectoryCrafter~\cite{yu2025trajectorycrafter} & 104.31 & 87.18 & 22.55 & 39.85 & 48.72 & 45.87 \\
ReDirector~\cite{park2025redirector} & 69.42 & 55.79 & 14.91 & 19.82 & 23.04 & 20.08 \\
GEN3C~\cite{ren2025gen3c} & 65.53 & 52.62 & 14.64 & 20.52 & 25.89 & 20.49 \\
DepthDirector~\cite{chen2026depthdirector} & 71.43 & 57.53 & 15.39 & 28.84 & 32.95 & 21.34 \\
Geo-Align~\cite{li2026geoalign} & 68.68 & 55.49 & 14.61 & 21.33 & 27.84 & 14.13 \\
FlashRender~\cite{park2026flashrender} & 62.86 & 50.13 & 13.91 & \underline{16.99} & \underline{17.24} & \underline{12.94} \\
NeoVerse~\cite{yang2026neoverse} & 86.07 & 70.19 & 19.21 & 35.27 & 49.23 & 32.21 \\
\ourmethod{} & \underline{55.66} & \textbf{44.84} & \underline{12.57} & \textbf{12.61} & \textbf{10.45} & 13.95 \\
\midrule
InSpatio-World~\cite{inspatio2026world} & 71.44 & 56.76 & 15.64 & 23.50 & 35.05 & 15.70 \\
\ourmethod{}-DMD & \textbf{49.56} & \textbf{39.73} & \textbf{10.85} & \textbf{20.25} & \textbf{34.65} & \textbf{6.05} \\
\bottomrule\end{tabular}}
\end{table}

\vspace{1mm}
\noindent\textbf{Results.}
On orbit cases where both methods pass the orientation check, the excess of \ourmethod{} is 8.41\,mm lower than that of GEN3C (139 pairs, 95\% paired bootstrap CI [$-$13.18, $-$3.85], $p<0.01$); per-method values are in \cref{tab:supp:motion_main}, and the paired comparison with GEN3C is in \cref{tab:supp:motion_significance}. The paired values (10.41 and 18.82\,mm) differ from the per-method values in \cref{tab:supp:motion_main} (13.95 and 20.49\,mm) because they use only cases where both methods pass the check. This is consistent with the body distortions of GEN3C in cases that pass the joint evaluation (\cref{fig:gen3c_pass_more}).
\ourmethod{}-DMD and ReCamMaster have the lowest MPJPE, but the body fails the orientation check in 34.65\% and 32.49\% of their orbit cases, compared with 10.45\% for \ourmethod{}.
On non-orbit and tilt trajectories, where no orientation check applies, \ourmethod{} reaches 46.88 and 52.57\,mm MPJPE; the lowest autoregressive values are 43.30 and 49.67\,mm (\ourmethod{}-DMD), and the lowest values overall are 42.80 (FlashRender) and 45.12\,mm (ReCamMaster; \cref{tab:supp:motion_other}).
The difference appears under orbits that need rotation. On orbit cases, \ourmethod{} is the only method in the top two for both body orientation and MPJPE (\cref{tab:supp:motion_main}). The other methods trade one measure for the other. A method that rotates the view too little keeps the body close to the source pose, so its MPJPE stays low but the body often does not turn with the camera: ReCamMaster has a similar MPJPE to \ourmethod{} but fails the orientation check in 32.49\% of cases (10.45\% for \ourmethod{}). A method that does reach the target side often distorts the body, which raises MPJPE: FlashRender has the next-lowest orientation error but a higher MPJPE (62.86 vs.\ 55.66\,mm).

\begin{table}[t]
\centering
\caption{\textbf{Motion preservation on non-orbit and tilt trajectories.} MPJPE and PA-MPJPE (mm) and ROT ($^\circ$), lower is better; no orientation check applies. Rows below the rule are autoregressive; best in \textbf{bold} and second \underline{underlined} within each group.}
\label{tab:supp:motion_other}
\footnotesize\setlength{\tabcolsep}{3pt}
\fitto{\columnwidth}{%
\begin{tabular}{lcccccc}\toprule
& \multicolumn{3}{c}{Non-orbit} & \multicolumn{3}{c}{Tilt} \\
\cmidrule(lr){2-4}\cmidrule(lr){5-7}
Method & MPJPE & PA & ROT & MPJPE & PA & ROT \\
\midrule
ReCamMaster~\cite{bai2025recammaster} & \underline{46.47} & 37.16 & 10.14 & \textbf{45.12} & \textbf{35.99} & \textbf{9.73} \\
TrajectoryCrafter~\cite{yu2025trajectorycrafter} & 73.63 & 59.35 & 16.19 & 77.44 & 63.42 & 16.89 \\
ReDirector~\cite{park2025redirector} & 50.35 & 39.85 & 10.85 & 62.51 & 50.01 & 13.80 \\
GEN3C~\cite{ren2025gen3c} & 53.01 & 41.10 & 11.58 & 66.13 & 53.10 & 14.77 \\
DepthDirector~\cite{chen2026depthdirector} & 52.27 & 41.33 & 11.41 & 59.12 & 45.74 & 13.44 \\
Geo-Align~\cite{li2026geoalign} & 50.01 & 40.10 & 10.71 & 62.27 & 49.24 & 13.77 \\
FlashRender~\cite{park2026flashrender} & \textbf{42.80} & \textbf{34.03} & \textbf{9.19} & 55.89 & 42.20 & 12.13 \\
NeoVerse~\cite{yang2026neoverse} & 58.29 & 45.93 & 12.93 & 73.42 & 58.50 & 16.10 \\
\ourmethod{} & 46.88 & \underline{37.05} & \underline{9.93} & \underline{52.57} & \underline{41.72} & \underline{11.39} \\
\midrule
InSpatio-World~\cite{inspatio2026world} & 54.45 & 42.84 & 11.67 & 61.68 & 49.21 & 13.35 \\
\ourmethod{}-DMD & \textbf{43.30} & \textbf{34.20} & \textbf{9.26} & \textbf{49.67} & \textbf{38.96} & \textbf{10.95} \\
\bottomrule\end{tabular}}
\end{table}

\begin{table*}[!t]
\centering
\caption{\textbf{Significance of the motion-preservation results.} \ourmethod{} (bidirectional) minus GEN3C. (a) Local-pose error between the source and the generated video on all trajectories of the camera-control benchmark (GVHMR, neutral body shape; MPJPE and PA-MPJPE in mm, ROT in degrees; negative favors \ourmethod{}). (b) Orbit cases: Excess is MPJPE minus the median viewpoint noise of \cref{tab:supp:motion_noise}, on cases where both methods pass the orientation check (negative favors \ourmethod{}). Only cases computable for both methods are paired. Brackets: 95\% percentile interval of a paired bootstrap that resamples source videos (10{,}000 replicates). $p$: two-sided exact sign test (ties dropped), not adjusted for multiplicity.}
\label{tab:supp:motion_significance}
\footnotesize\setlength{\tabcolsep}{3pt}
(a) Local-pose error, all trajectories\\[1mm]
\fitto{\textwidth}{%
\begin{tabular}{llcccc}\toprule
Split & \ourmethod{} vs.\ & Metric & Ours / other & $\Delta$ [95\% CI] & $p$ \\
\midrule
Standard & GEN3C & MPJPE (mm) & 46.47 / 54.86 & $-8.40$ [$-13.55$, $-4.26$] & $<$0.001 \\
 &  & PA-MPJPE (mm) & 37.08 / 43.00 & $-5.91$ [$-10.43$, $-2.29$] & $<$0.001 \\
 &  & ROT ($^\circ$) & 10.15 / 12.21 & $-2.06$ [$-2.99$, $-1.26$] & $<$0.001 \\
\midrule
Extended & GEN3C & MPJPE (mm) & 59.65 / 73.50 & $-13.85$ [$-20.52$, $-7.78$] & $<$0.001 \\
 &  & PA-MPJPE (mm) & 47.67 / 59.28 & $-11.62$ [$-17.73$, $-6.08$] & $<$0.001 \\
 &  & ROT ($^\circ$) & 13.17 / 16.09 & $-2.92$ [$-4.29$, $-1.79$] & $<$0.001 \\
\midrule
Combined & GEN3C & MPJPE (mm) & 49.89 / 59.70 & $-9.81$ [$-14.30$, $-5.93$] & $<$0.001 \\
 &  & PA-MPJPE (mm) & 39.83 / 47.23 & $-7.39$ [$-11.38$, $-4.06$] & $<$0.001 \\
 &  & ROT ($^\circ$) & 10.93 / 13.22 & $-2.28$ [$-3.08$, $-1.58$] & $<$0.001 \\
\bottomrule
\end{tabular}}

\vspace{2mm}
(b) Orbit cases\\[1mm]
\fitto{0.8\textwidth}{%
\begin{tabular}{llccc}\toprule
Measure & \ourmethod{} vs.\ & Ours / other & $\Delta$ [95\% CI] & $p$ \\
\midrule
Excess (mm) & GEN3C & 10.41 / 18.82 & $-8.41$ [$-13.18$, $-3.85$] & 0.003 \\
\bottomrule
\end{tabular}}
\end{table*}

\section{Single-image novel-view synthesis}
\label{sec:supp:single_image}
\begin{figure*}[t]
\centering
\includegraphics[width=\linewidth,height=.55\textheight,keepaspectratio]{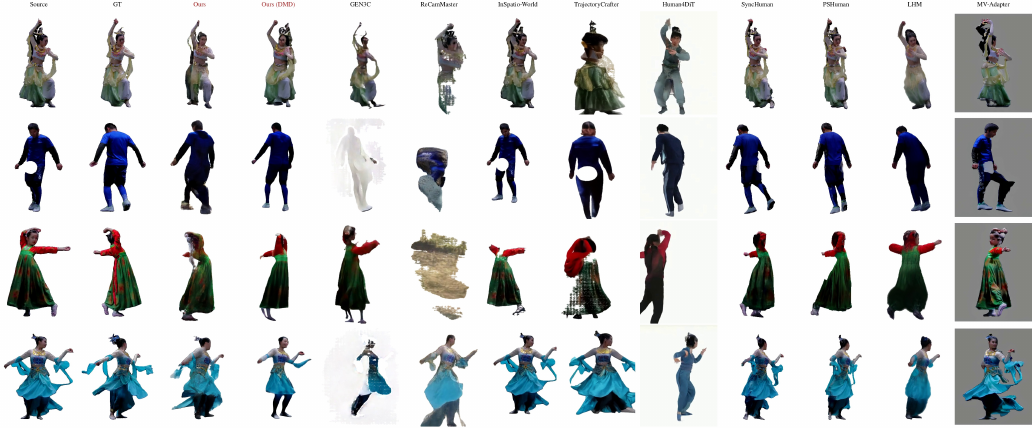}
\caption{\textbf{Human novel-view synthesis from a single source image.} As the camera circles the subject, each method must infer clothing and body surfaces hidden in the input view.}
\label{fig:qual:image}
\end{figure*}

This variant of \cref{sec:experiments:reconstruction} uses the same 12 subjects and orbits. The input is the first frame of the source window, repeated as a still video for video-based methods, and every target is captured at that same instant. MV-Adapter~\cite{huangmvadapter} produces fixed orthographic, box-aligned views; its orientation error and IoU are N/A, and its pixel scores are reported without ranking. \ourmethod{} (bidirectional) achieves 20.68\,dB PSNR, 0.678 IoU, and $7.7^\circ$ mean orientation error (\cref{tab:exp:reconstruction_image}), compared with $46.3^\circ$ for GEN3C, $58.4^\circ$ for ReCamMaster, and $71.2^\circ$ for TrajectoryCrafter; \ourmethod{}-DMD reaches $10.2^\circ$ versus $72.3^\circ$ for InSpatio-World~\cite{inspatio2026world}. PSHuman~\cite{lipshuman} and SyncHuman~\cite{chensynchuman} retain stronger mask overlap and DINO similarity, and \ourmethod{}-DMD obtains the lowest LPIPS. \Cref{fig:qual:image} shows qualitative results.

\begin{table}[!t]
\centering
\caption{\textbf{Single-image novel-view synthesis.} Rows below the rule are autoregressive. Gray values are unranked. $^{\dagger}$Explicit 3D methods render at target cameras, which constrains orientation by construction. MV-Adapter uses fixed orthographic views: orientation and IoU are N/A; its pixel scores ($^{\S}$) use a different canvas and are unranked. LA (large-angle orientation success) measures the fraction of eligible keyframes within $30^\circ$ of the target orientation when both the azimuth change and GT rotation reach $30^\circ$. Missing subjects count as failures. TrajectoryCrafter generates 49 frames.}
\label{tab:exp:reconstruction_image}
\footnotesize\setlength{\tabcolsep}{3pt}
\fitto{\columnwidth}{%
\begin{tabular}{lccccccc}\toprule
Method & Orient.$\downarrow$ & LA$\uparrow$ & IoU$\uparrow$ & DINO$\uparrow$ & PSNR$\uparrow$ & SSIM$\uparrow$ & LPIPS$\downarrow$ \\
\midrule
Human4DiT~\cite{shaohuman4dit} & 11.1 & 91 & 0.356 & 0.555 & 17.06 & 0.925 & 0.114 \\
SyncHuman~\cite{chensynchuman}$^{\dagger}$ & \textbf{5.5} & \textbf{99} & \underline{0.694} & \textbf{0.817} & 20.14 & \underline{0.933} & \underline{0.070} \\
PSHuman~\cite{lipshuman}$^{\dagger}$ & \underline{6.0} & \underline{98} & \textbf{0.709} & \underline{0.812} & \underline{20.57} & \textbf{0.934} & \underline{0.070} \\
LHM~\cite{qiulhm}$^{\dagger}$ & 11.7 & 91 & 0.623 & 0.711 & 19.65 & \textbf{0.934} & 0.080 \\
MV-Adapter~\cite{huangmvadapter} & N/A & N/A & N/A & 0.780 & \unr{16.67\tnmark{\S}} & \unr{0.686\tnmark{\S}} & \unr{0.237\tnmark{\S}} \\
GEN3C~\cite{ren2025gen3c} & 46.3 & 37 & 0.526 & 0.606 & 18.13 & 0.924 & 0.095 \\
ReCamMaster~\cite{bai2025recammaster} & 58.4 & 19 & 0.327 & 0.634 & 18.94 & 0.930 & 0.080 \\
TrajectoryCrafter~\cite{yu2025trajectorycrafter} & 71.2 & 15 & 0.399 & 0.611 & 19.93 & \underline{0.933} & 0.075 \\
\ourmethod{} & 7.7 & 97 & 0.678 & 0.760 & \textbf{20.68} & \underline{0.933} & \textbf{0.066} \\
\midrule
InSpatio-World~\cite{inspatio2026world} & 72.3 & 8 & 0.540 & \textbf{0.781} & 19.19 & 0.929 & 0.079 \\
\ourmethod{}-DMD & \textbf{10.2} & \textbf{95} & \textbf{0.632} & 0.752 & \textbf{20.23} & \textbf{0.934} & \textbf{0.062} \\
\bottomrule\end{tabular}}
\end{table}

\section{Motion-control evaluation details}
\label{sec:supp:motion50}
\begin{figure}[t]
\centering
\begin{minipage}[b]{\linewidth}\centering
\includegraphics[width=\linewidth,height=.34\textheight,keepaspectratio]{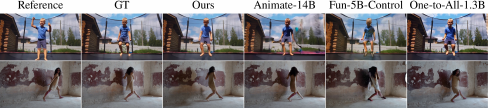}\\[-1pt]
{\small (a) Landscape}
\end{minipage}\par\smallskip
\begin{minipage}[b]{\linewidth}\centering
\includegraphics[width=\linewidth,height=.34\textheight,keepaspectratio]{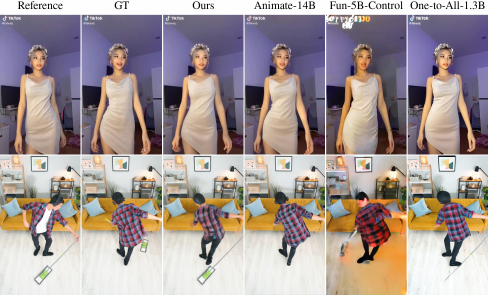}\\[-1pt]
{\small (b) Portrait}
\end{minipage}
\caption{\textbf{Motion control from a reference image in landscape and portrait videos.} The examples show how closely each output follows the driving poses and retains the reference subject's appearance as the body moves.}
\label{fig:motion_formats}
\end{figure}

The 50 cases comprise 24 RealisDance-Val (RDV) clips (9 static-camera and 15 with weak camera motion), 18 UBC Fashion~\cite{zablotskaia2019dwnet} clips, and 8 TikTok~\cite{jafarian2021tiktok} clips, all using frames 0--80 with frame 0 as the reference (\cref{tab:supp:motion_datasets}). TikTok selections exclude the published training IDs of One-to-All-1.3B~\cite{shi2026onetoall}. Wan2.2-Animate-14B~\cite{cheng2025wananimate} receives per-frame ground-truth face crops in addition to pose control; the other methods receive no ground-truth RGB beyond the reference image. \Cref{fig:motion_formats} shows landscape and portrait examples. Portrait clips ($480\times832$) are padded by edge replication to \ourmethod{}'s landscape resolution and cropped back after generation, and watermarks are masked identically in predictions and ground truth.

\begin{table*}[t]
\centering
\caption{\textbf{Motion-control results by dataset.} $^{\dagger}$Unranked on UBC because of reported training overlap. $^{\ddagger}$Wan2.2-Animate-14B receives ground-truth face crops, so its ArcFace is unranked.}
\label{tab:supp:motion_datasets}
\small\setlength{\tabcolsep}{4pt}
\fitto{\textwidth}{%
\begin{tabular}{llcccccccc}\toprule
Dataset & Method & PSNR$\uparrow$ & SSIM$\uparrow$ & LPIPS$\downarrow$ & ArcFace$\uparrow$ & PCK@0.1$\uparrow$ & PCK@0.05$\uparrow$ & FID$\downarrow$ & FVD$\downarrow$ \\
\midrule
RDV (24) & Wan2.2-Animate-14B~\cite{cheng2025wananimate} & \underline{20.06} & \underline{0.624} & \underline{0.209} & \unr{0.321\tnmark{\ddagger}} & \underline{0.919} & \textbf{0.867} & \textbf{38.7} & 372 \\
 & Wan2.2-Fun-5B-Control~\cite{alibabapai2025wanfun} & 17.13 & 0.542 & 0.299 & 0.152 & 0.844 & 0.749 & 64.6 & 624 \\
 & One-to-All-1.3B~\cite{shi2026onetoall} & 17.67 & 0.570 & 0.272 & \textbf{0.243} & 0.871 & 0.730 & 53.2 & 580 \\
\cmidrule{2-10}
 & \ourmethod{} & \textbf{20.89} & \textbf{0.692} & \textbf{0.203} & \underline{0.241} & \textbf{0.921} & \underline{0.851} & \underline{51.5} & 608 \\
\midrule
UBC (18) & Wan2.2-Animate-14B~\cite{cheng2025wananimate} & \textbf{25.07} & \textbf{0.908} & \textbf{0.054} & \unr{0.640\tnmark{\ddagger}} & \textbf{0.997} & \textbf{0.995} & \textbf{18.3} & 66 \\
 & Wan2.2-Fun-5B-Control~\cite{alibabapai2025wanfun} & 23.04 & 0.889 & 0.083 & \textbf{0.358} & \underline{0.996} & 0.993 & 31.3 & 123 \\
 & One-to-All-1.3B~\cite{shi2026onetoall}$^{\dagger}$ & \unr{18.48} & \unr{0.845} & \unr{0.115} & \unr{0.478} & \unr{0.997} & \unr{0.994} & \unr{29.2} & \unr{148} \\
\cmidrule{2-10}
 & \ourmethod{} & \underline{24.95} & \underline{0.901} & \underline{0.074} & \underline{0.327} & \textbf{0.997} & \underline{0.994} & \underline{23.2} & 78 \\
\midrule
TikTok (8) & Wan2.2-Animate-14B~\cite{cheng2025wananimate} & \underline{21.48} & 0.817 & \textbf{0.175} & \unr{0.461\tnmark{\ddagger}} & \underline{0.980} & \textbf{0.952} & \textbf{48.2} & 279 \\
 & Wan2.2-Fun-5B-Control~\cite{alibabapai2025wanfun} & 18.60 & 0.758 & 0.275 & 0.265 & \underline{0.980} & 0.936 & 80.8 & 471 \\
 & One-to-All-1.3B~\cite{shi2026onetoall} & 20.38 & \underline{0.840} & \underline{0.177} & \textbf{0.432} & 0.978 & 0.926 & 72.9 & 303 \\
\cmidrule{2-10}
 & \ourmethod{} & \textbf{22.02} & \textbf{0.845} & 0.193 & \underline{0.275} & \textbf{0.984} & \underline{0.950} & \underline{68.1} & 338 \\
\bottomrule\end{tabular}}
\end{table*}

\vspace{1mm}
\noindent\textbf{Metrics.}
Full-frame PSNR, SSIM, and LPIPS are averaged within each case and then across cases. ArcFace~\cite{deng2019arcface} compares generated and ground-truth faces at the same frame where both are detected. PCK uses Keypoint R-CNN with thresholds of 0.1 and 0.05 times the person-box long side. FID and FVD are recomputed from the pooled outputs of each reported dataset; FVD uses five 16-frame clips per video and is a small-sample estimate. Excluding the reference frame changes PSNR/SSIM/LPIPS by at most 0.15/0.002/0.002 and leaves the ranking on every metric unchanged.

\vspace{1mm}
\noindent\textbf{Per-case comparisons.}
Against Wan2.2-Animate-14B, \ourmethod{} wins PSNR on 29/50 cases, SSIM on 23/50, LPIPS on 15/50, and ArcFace on 4/47; against Wan2.2-Fun-5B-Control~\cite{alibabapai2025wanfun}, 48/50, 46/50, 38/50, and 29/47; against One-to-All-1.3B, 49/50, 49/50, 45/50, and 13/47.

\section{Joint-control visual quality}
\label{sec:supp:joint_quality}
\ourmethod{} has the lowest six-dimension VBench mean (0.813 vs.\ 0.824--0.831), mainly from aesthetic and imaging quality (\cref{tab:supp:joint_vbench}).

\begin{table}[!t]
\centering
\caption{\textbf{VBench scores for joint camera and motion control.} Camera movement can reduce subject/background consistency and temporal-flickering scores; the scores should be interpreted together with camera and human-control accuracy.}
\label{tab:supp:joint_vbench}
\footnotesize\setlength{\tabcolsep}{3pt}
\fitto{\columnwidth}{%
\begin{tabular}{lccccccc}\toprule
& & \multicolumn{6}{c}{VBench dimensions} \\
\cmidrule(lr){3-8}
Method & Overall & Subj. & Bg. & Aesth. & Img. & Flicker & Smooth. \\
\midrule
CamAnimate~\cite{wang2024humanvid} & \underline{0.826} & \underline{0.904} & \textbf{0.931} & \textbf{0.544} & 0.644 & 0.957 & 0.976 \\
RealisDance-DiT~\cite{zhou2025realisdance} & \textbf{0.831} & \textbf{0.912} & \underline{0.929} & 0.538 & \textbf{0.651} & \textbf{0.970} & \textbf{0.988} \\
Uni3C~\cite{cao2025uni3c} & 0.824 & 0.891 & 0.917 & \underline{0.542} & \underline{0.647} & 0.959 & \underline{0.986} \\
\midrule
\ourmethod{} & 0.813 & 0.881 & 0.926 & 0.504 & 0.609 & \underline{0.968} & \textbf{0.988} \\
\bottomrule\end{tabular}}
\end{table}

\begin{figure*}[p]
\centering
\includegraphics[width=\linewidth,height=.85\textheight,keepaspectratio]{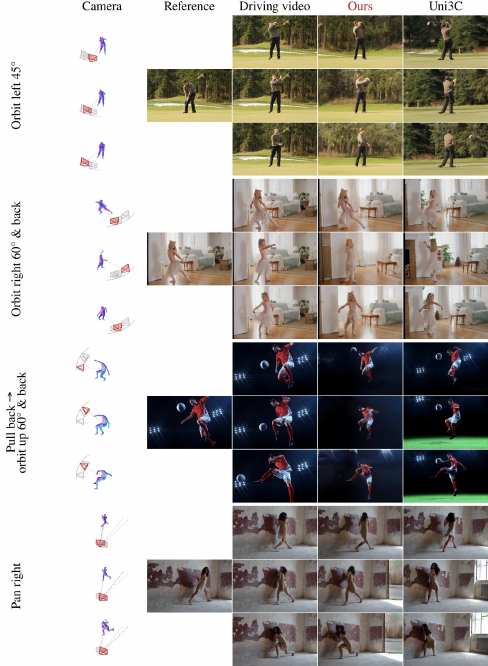}
\caption{\textbf{Additional joint camera and motion control cases, \ourmethod{} versus Uni3C.} This extends \cref{fig:joint_control_visual}. Background features indicate camera motion, while the subject's pose and visible side show whether the driving motion and orientation are preserved.}
\label{fig:joint_control_all}
\end{figure*}

\Cref{fig:joint_control_all} adds joint camera and motion control cases to \cref{fig:joint_control_visual}. In these cases, \ourmethod{} shows the person from the commanded viewpoint more consistently than Uni3C, and its poses follow the driving video more closely.

\section{Long-video camera control}
\label{sec:supp:long_video}
\begin{figure*}[t]
\centering
\includegraphics[width=\linewidth]{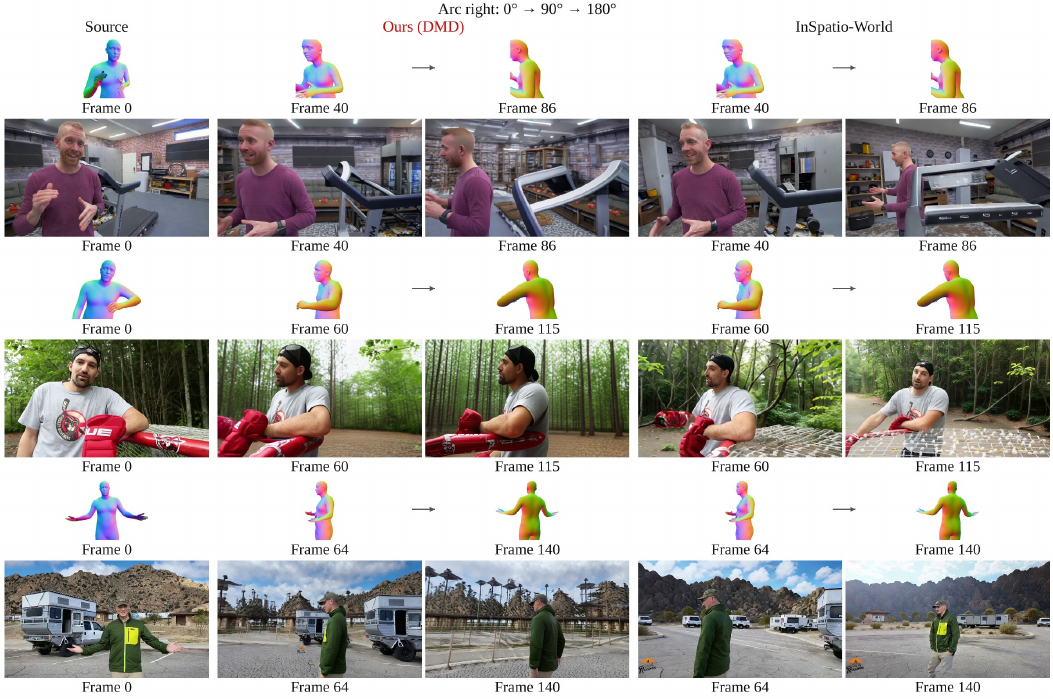}
\caption{\textbf{Long-video camera control.} Each row pair shows the source at frame 0 and two later frames from \ourmethod{}-DMD and InSpatio-World, with the body-mesh rendering of the commanded view above each frame.}
\label{fig:supp:long_video}
\end{figure*}

We evaluate long-horizon generation with the causal model on ten source videos from the camera-control set, each paired with five camera trajectories (50 cases). We compare \ourmethod{}-DMD with InSpatio-World~\cite{inspatio2026world}, the other autoregressive method. Long videos have 162 frames, scored as two consecutive 81-frame windows (frames 0--80 and 81--161). Window metrics average the two windows, and whole-video quality is computed separately. The five camera trajectories are: (i) a continuous right arc of $90^\circ$ per window ($0^\circ \to 90^\circ \to 180^\circ$); (ii) a continuous right arc of $45^\circ$ per window ($0^\circ \to 45^\circ \to 90^\circ$); (iii) an upward arc-and-return ($0^\circ \to 60^\circ \to 0^\circ$); (iv) tilt up--down--return; and (v) zoom in--out--return, with camera-to-subject distance ratios $1 \to 0.5 \to 1.5 \to 1$. The last three trajectories repeat in each window. Camera errors are estimator outputs rather than recovered ground-truth cameras, and position and size use fixed-depth planar proxies. DINO measures appearance consistency within each video and is averaged over the two windows, and MEt3R compares each generation with its source rather than with ground-truth novel views. On this set, \ourmethod{}-DMD has lower body-orientation, position, size, and missing-subject errors and higher DINO consistency, whereas InSpatio-World has lower ViPE rotation error (\cref{tab:supp:long_video}); \cref{fig:supp:long_video} shows examples. Visual-quality scores are mixed, and we do not claim an overall winner.

\begin{table}[t]
\centering
\caption{\textbf{Long-video camera control.} Window metrics average two consecutive 81-frame windows (frames 0--80 and 81--161); GVHMR orientation averages the 79 windows valid for both methods. Position and size are fixed-depth planar proxies. MEt3R is computed against the source video; restricting it to frame pairs valid for both methods gives the same ranking (0.280 vs.\ 0.313).}
\label{tab:supp:long_video}
\footnotesize\setlength{\tabcolsep}{4pt}
\fitto{\columnwidth}{%
\begin{tabular}{lcc}\toprule
Metric & \ourmethod{}-DMD & InSpatio-World~\cite{inspatio2026world} \\
\midrule
\multicolumn{3}{l}{\textit{Camera and subject (window mean)}} \\
ViPE rotation error ($^\circ$)$\downarrow$ & 11.22 & \textbf{7.16} \\
GVHMR orientation error ($^\circ$)$\downarrow$ & \textbf{14.48} & 15.56 \\
Position proxy error$\downarrow$ & \textbf{0.046} & 0.062 \\
Size log proxy error$\downarrow$ & \textbf{0.205} & 0.260 \\
Missing rate (\%)$\downarrow$ & \textbf{0.06} & 0.37 \\
DINO$\uparrow$ & \textbf{0.884} & 0.866 \\
\midrule
\multicolumn{3}{l}{\textit{Whole-video quality}} \\
MEt3R$\downarrow$ & \textbf{0.283} & 0.314 \\
VBench Subj.$\uparrow$ & \textbf{0.895} & 0.885 \\
VBench Bg.$\uparrow$ & 0.911 & \textbf{0.914} \\
VBench Smooth.$\uparrow$ & \textbf{0.981} & 0.972 \\
VBench Aesth.$\uparrow$ & 0.606 & \textbf{0.608} \\
VBench Img.$\uparrow$ & 0.740 & \textbf{0.752} \\
\bottomrule\end{tabular}}
\end{table}

\vspace{1mm}
\noindent\textbf{Generation speed.}
\Cref{tab:supp:speed} reports the time to generate an 81-frame $832\times480$ clip. Timing covers denoising only and excludes model loading and VAE encoding and decoding, for both models. The bidirectional model uses 50 sampling steps; \ourmethod{}-DMD uses four denoising steps per chunk. On an RTX 6000 Ada, \ourmethod{}-DMD takes about twice the clip's playback duration, so it is not real-time.

\begin{table}[t]
\centering
\caption{\textbf{Generation speed for an 81-frame clip.} Timing covers denoising only, excluding model loading and VAE encoding and decoding. \ourmethod{}-DMD times are the mean $\pm$ sample standard deviation over three runs (medians: 21.38\,s on RTX 3090, 10.70\,s on RTX 6000 Ada). Ratio: generation time divided by the clip's playback duration (81 frames at 15\,fps, 5.4\,s). Frames/s and Ratio use the median time. Memory: peak allocated during rollout / peak whole-GPU usage.}
\label{tab:supp:speed}
\footnotesize\setlength{\tabcolsep}{3pt}
\fitto{\columnwidth}{%
\begin{tabular}{llcccc}\toprule
Model & GPU & Time (s)$\downarrow$ & Frames/s$\uparrow$ & Ratio$\downarrow$ & Memory (GiB) \\
\midrule
\ourmethod{} (bidirectional) & RTX 6000 Ada & 122 & 0.66 & 22.6$\times$ & -- \\
\midrule
\ourmethod{}-DMD & RTX 3090 & 21.43 $\pm$ 0.51 & 3.79 & 3.96$\times$ & 16.28 / 17.22 \\
\ourmethod{}-DMD & RTX 6000 Ada & 10.70 $\pm$ 0.22 & 7.57 & 1.98$\times$ & 16.28 / 17.41 \\
\bottomrule\end{tabular}}
\end{table}

\section{Conditioning ablation}
\label{sec:supp:ablation}
\begin{figure}[t]
\centering
\includegraphics[width=\linewidth]{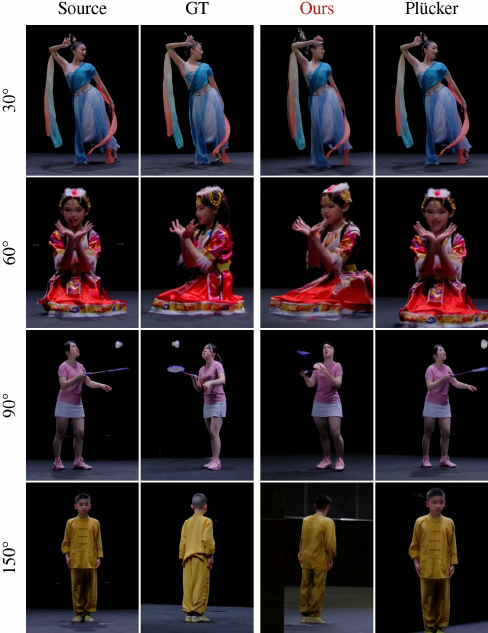}
\caption{\textbf{Qualitative conditioning ablation on multi-view captures.} PKU-DyMVHumans subjects at target-camera separations of $30^\circ$, $60^\circ$, $90^\circ$, and $150^\circ$ (rows), showing the source view, the ground-truth target view, \ourmethod{} (Ours), and the Plücker variant. This corresponds to the top part of \cref{tab:supp:ablation_nvs}.}
\label{fig:supp:ablation_nvs_visual}
\end{figure}

\begin{figure*}[t]
\centering
{\raggedright
\includegraphics[scale=1.91,trim=0 0 236.3 0,clip]{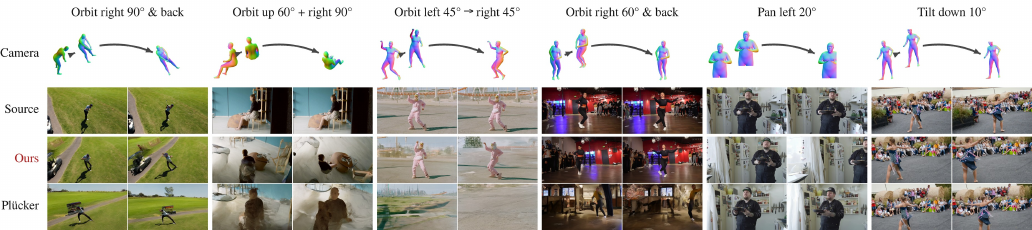}\\[3pt]
\includegraphics[scale=1.91,trim=0 0 474.6 0,clip]{figures/visual_results/fig_ablation_plucker_q92.pdf}%
\includegraphics[scale=1.91,trim=258.7 0 0 0,clip]{figures/visual_results/fig_ablation_plucker_q92.pdf}\par}
\caption{\textbf{Qualitative conditioning ablation.} Camera-control cases with the trajectory named above each column; the top row illustrates the commanded camera motion. Each case shows two frames of the source video, \ourmethod{} (Ours, body-mesh conditioning), and the Plücker variant.}
\label{fig:supp:ablation_visual}
\end{figure*}

We ablate the conditioning representation at an early training stage because later training adds motion-control data, which camera rays cannot express; at 10k steps, the body-mesh model already performs comparably to the final model on camera control (joint success 81.03\% vs.\ 78.62\%, orientation error 13.06$^\circ$ vs.\ 12.61$^\circ$; the final checkpoint is used because it also supports motion control). The Plücker variant replaces the body-mesh renderings with per-pixel Plücker maps~\cite{he2024cameractrl} of the source and target cameras and keeps the backbone, source--target concatenation, and batch size. Both variants are trained for 10k steps with the same recipe and training data. In \cref{tab:supp:ablation_quality,tab:supp:ablation_nvs,tab:supp:ablation_image,tab:supp:ablation}, \ourmethod{} denotes the body-mesh model and Plücker the variant with per-pixel Plücker maps, fine-tuned from the same backbone; the tables follow the visual-quality, novel-view synthesis, single-image, and camera-control tables, respectively. \Cref{fig:supp:ablation_nvs_visual,fig:supp:ablation_visual} show qualitative comparisons on multi-view captures and on in-the-wild camera control. The body-mesh model leads on every camera-control and subject-placement metric, by large margins on joint success (81.03\% vs.\ 44.83\%) and body orientation. Visual quality is comparable. On novel-view synthesis, it is better on orientation and silhouette IoU in every setting; Plücker is slightly better on DINO for multi-view captures, on all three pixel metrics for background-free $360^\circ$ video inputs, and on SSIM and LPIPS for single-image inputs. In the $360^\circ$ and single-image settings, Plücker rarely shows the requested side of the body (orientation error 76--82$^\circ$, 7\% LA for single images), and pixel metrics do not reflect this failure.

\begin{table}[t]
\centering
\caption{\textbf{Conditioning ablation on visual quality (in-the-wild human videos).} Columns follow \cref{tab:supp:quality}; Overall is the mean of the five VBench dimensions.}
\label{tab:supp:ablation_quality}
\footnotesize\setlength{\tabcolsep}{2pt}
\fitto{\columnwidth}{%
\begin{tabular}{lccccccccc}\toprule
& & \multicolumn{6}{c}{VBench$\uparrow$} & & \\
\cmidrule(lr){3-8}
Method & MEt3R$\downarrow$ & Overall & Subj. & Bg. & Smooth. & Aesth. & Img. & FID$\downarrow$ & FVD$\downarrow$ \\
\midrule
\ourmethod{} & \textbf{0.263} & \textbf{0.817} & \textbf{0.903} & 0.923 & \textbf{0.988} & \textbf{0.578} & \textbf{0.692} & \textbf{58.0} & \textbf{818} \\
Plücker & 0.266 & 0.816 & 0.900 & \textbf{0.925} & \textbf{0.988} & 0.577 & 0.689 & 59.5 & \textbf{818} \\
\bottomrule\end{tabular}}
\end{table}

\begin{table}[t]
\centering
\caption{\textbf{Conditioning ablation on novel-view synthesis (PKU-DyMVHumans).} Protocols follow \cref{tab:exp:nvs}. Top: multi-view captures with calibrated target cameras. Bottom: $360^\circ$ rendering from a background-free video input.}
\label{tab:supp:ablation_nvs}
\footnotesize\setlength{\tabcolsep}{3pt}
\fitto{\columnwidth}{%
\begin{tabular}{lcccccc}\toprule
Method & Orient.$\downarrow$ & IoU$\uparrow$ & DINO$\uparrow$ & PSNR$\uparrow$ & SSIM$\uparrow$ & LPIPS$\downarrow$ \\
\midrule
\multicolumn{7}{l}{\textit{Multi-view captures}} \\
\ourmethod{} & \textbf{32.1} & \textbf{0.550} & 0.752 & \textbf{24.75} & \textbf{0.951} & \textbf{0.031} \\
Plücker & 71.7 & 0.528 & \textbf{0.782} & 24.53 & 0.950 & 0.033 \\
\midrule
\multicolumn{7}{l}{\textit{$360^\circ$ rendering, video input}} \\
\ourmethod{} & \textbf{22.4} & \textbf{0.574} & \textbf{0.716} & 18.74 & 0.931 & 0.083 \\
Plücker & 81.8 & 0.230 & 0.660 & \textbf{19.22} & \textbf{0.933} & \textbf{0.074} \\
\bottomrule\end{tabular}}
\end{table}

\begin{table}[t]
\centering
\caption{\textbf{Conditioning ablation on single-image novel-view synthesis.} Protocol and metrics follow \cref{tab:exp:reconstruction_image}.}
\label{tab:supp:ablation_image}
\footnotesize\setlength{\tabcolsep}{3pt}
\fitto{\columnwidth}{%
\begin{tabular}{lccccccc}\toprule
Method & Orient.$\downarrow$ & LA$\uparrow$ & IoU$\uparrow$ & DINO$\uparrow$ & PSNR$\uparrow$ & SSIM$\uparrow$ & LPIPS$\downarrow$ \\
\midrule
\ourmethod{} & \textbf{10.8} & \textbf{93} & \textbf{0.630} & \textbf{0.685} & \textbf{18.97} & 0.930 & 0.081 \\
Plücker & 76.1 & 7 & 0.153 & 0.455 & 18.84 & \textbf{0.931} & \textbf{0.080} \\
\bottomrule\end{tabular}}
\end{table}

\begin{table*}[t]
\centering
\caption{\textbf{Conditioning ablation on camera control.} Body-mesh conditioning versus per-pixel Plücker maps. Columns and protocol follow \cref{tab:exp:camera} (580 cases; orientation and F/S/B from tracking-based GVHMR).}
\label{tab:supp:ablation}
\small\setlength{\tabcolsep}{4pt}\renewcommand{\arraystretch}{0.88}\setlength{\aboverulesep}{0.8pt}\setlength{\belowrulesep}{0.8pt}\setlength{\cmidrulesep}{0pt}
\fitto{\textwidth}{%
\begin{tabular}{lcccccccc}\toprule
& & \multicolumn{2}{c}{Background (ViPE)} & \multicolumn{2}{c}{Human (GVHMR)} & \multicolumn{3}{c}{Subject placement} \\
\cmidrule(lr){3-4}\cmidrule(lr){5-6}\cmidrule(lr){7-9}
Method & Joint (\%)$\uparrow$ & Rot.$\downarrow$ & Trans.$\downarrow$ & Orient.$\downarrow$ & F/S/B (\%)$\uparrow$ & Pos.\ ($/H$)$\downarrow$ & Miss (\%)$\downarrow$ & Size err.$\downarrow$ \\
\midrule
\ourmethod{} & \textbf{81.03} & \textbf{7.64} & \textbf{0.034} & \textbf{13.06} & \textbf{85.80} & \textbf{0.060} & \textbf{2.8} & \textbf{0.090} \\
Plücker & 44.83 & 10.77 & 0.055 & 18.56 & 80.19 & 0.157 & 5.4 & 0.115 \\
\bottomrule\end{tabular}}
\end{table*}

\section{Discussion: in-the-wild use}
\label{sec:supp:discussion}
The examples in this section are qualitative and lie outside the benchmarks; we did not measure the realized camera angles.

\begin{figure*}[t]
\centering
\includegraphics[width=\linewidth]{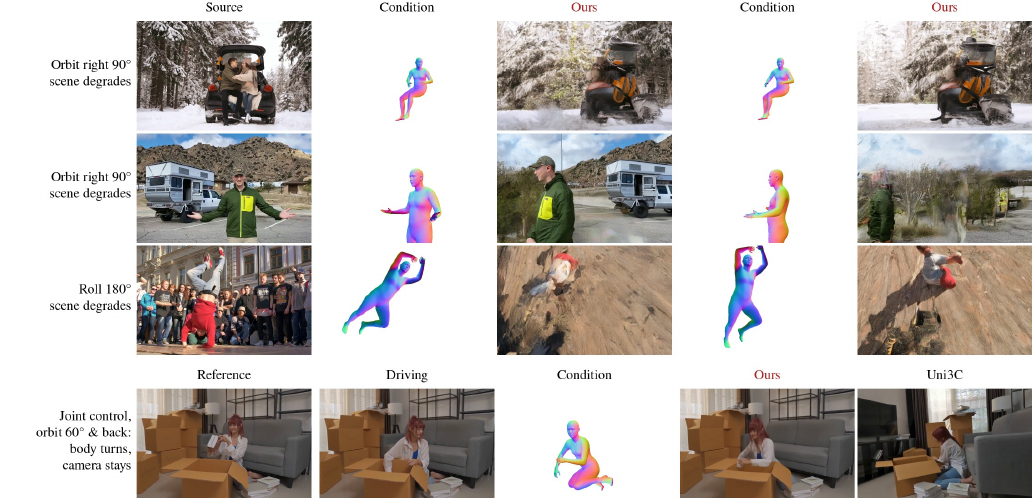}
\caption{\textbf{Failure cases.} Top: camera control; regions the source does not show can degrade, sometimes progressively over the clip. Bottom: joint camera and motion control on a seated, heavily occluded person at the frame of the largest commanded rotation ($60^\circ$): \ourmethod{} turns the body while the background stays as in the reference and driving frames.}
\label{fig:supp:failures}
\end{figure*}

\vspace{1mm}
\noindent\textbf{Failure cases.}
\Cref{fig:supp:failures} shows typical failures. Under fast or large-angle orbits, regions of the scene that the source video does not show can degrade. Most other camera-control methods degrade in these cases as well, since these regions must be synthesized without observation. In joint camera and motion control, when the person is heavily occluded or only a few body parts are visible, the condition may not tell whether the camera or the person is moving. For example, with only the upper body in view, a person bending forward and an upward camera arc can produce nearly the same condition, so the model cannot tell them apart; in the bottom example of \cref{fig:supp:failures}, \ourmethod{} turns the body instead of moving the camera.

\begin{figure*}[t]
\centering
\makebox[\linewidth][c]{\includegraphics[width=.94\paperwidth,height=.86\textheight,keepaspectratio]{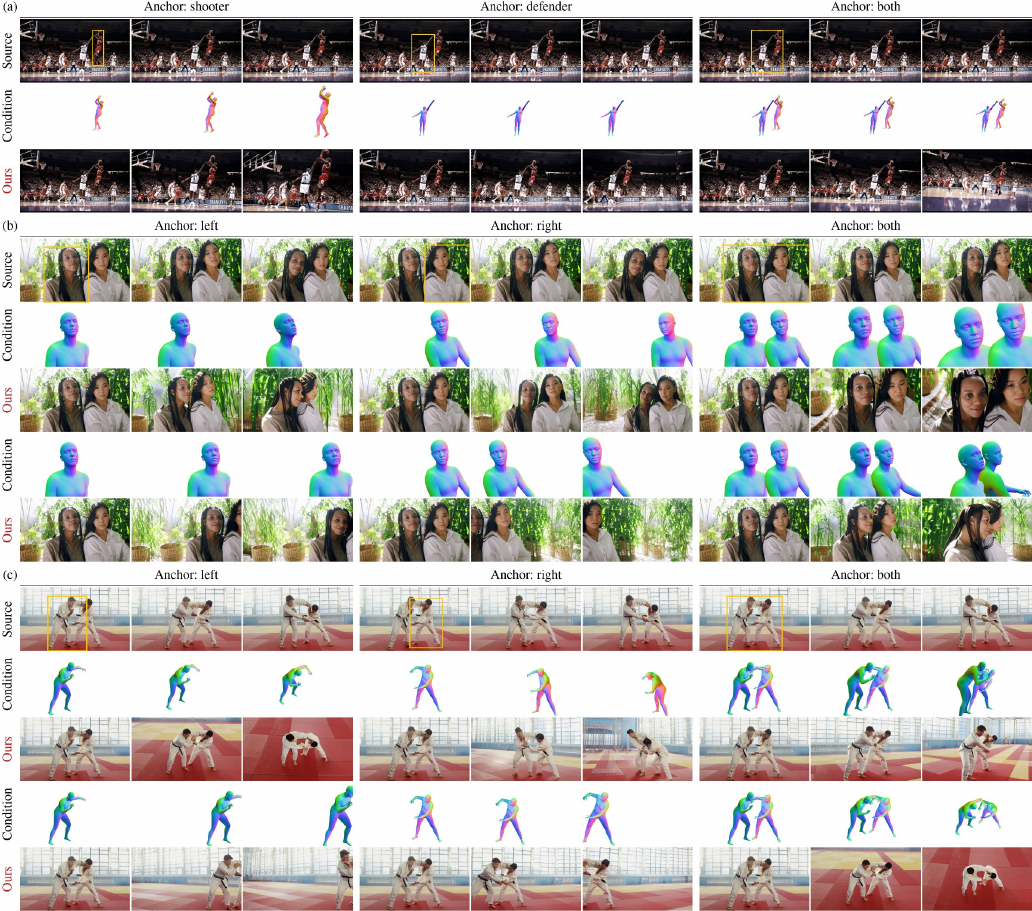}}
\caption{\textbf{Anchor selection in a multi-person scene.} For each scene, the same source clip is generated with different single people or a group selected as the anchor.}
\label{fig:supp:anchor_choice}
\end{figure*}

\vspace{1mm}
\noindent\textbf{Choosing the anchor in a multi-person scene.}
The user selects the anchor by passing person boxes on the first source frame to SAM 3D Body, so in a scene with several people the camera can be defined around one person or a group. The selected people are reconstructed as meshes with fixed poses and rendered together along the camera trajectory, giving one normal-map condition sequence. The camera orbits the center of the meshes' combined axis-aligned 3D bounding box, with a constant radius and focal length within each orbit. People who are not selected appear only in the source video and are not constrained by the condition. \Cref{fig:supp:anchor_choice} shows target orbits generated with different single people and with multiple people as the anchor.

\begin{figure*}[t]
\centering
\makebox[\linewidth][c]{\includegraphics[width=.94\paperwidth,height=.86\textheight,keepaspectratio]{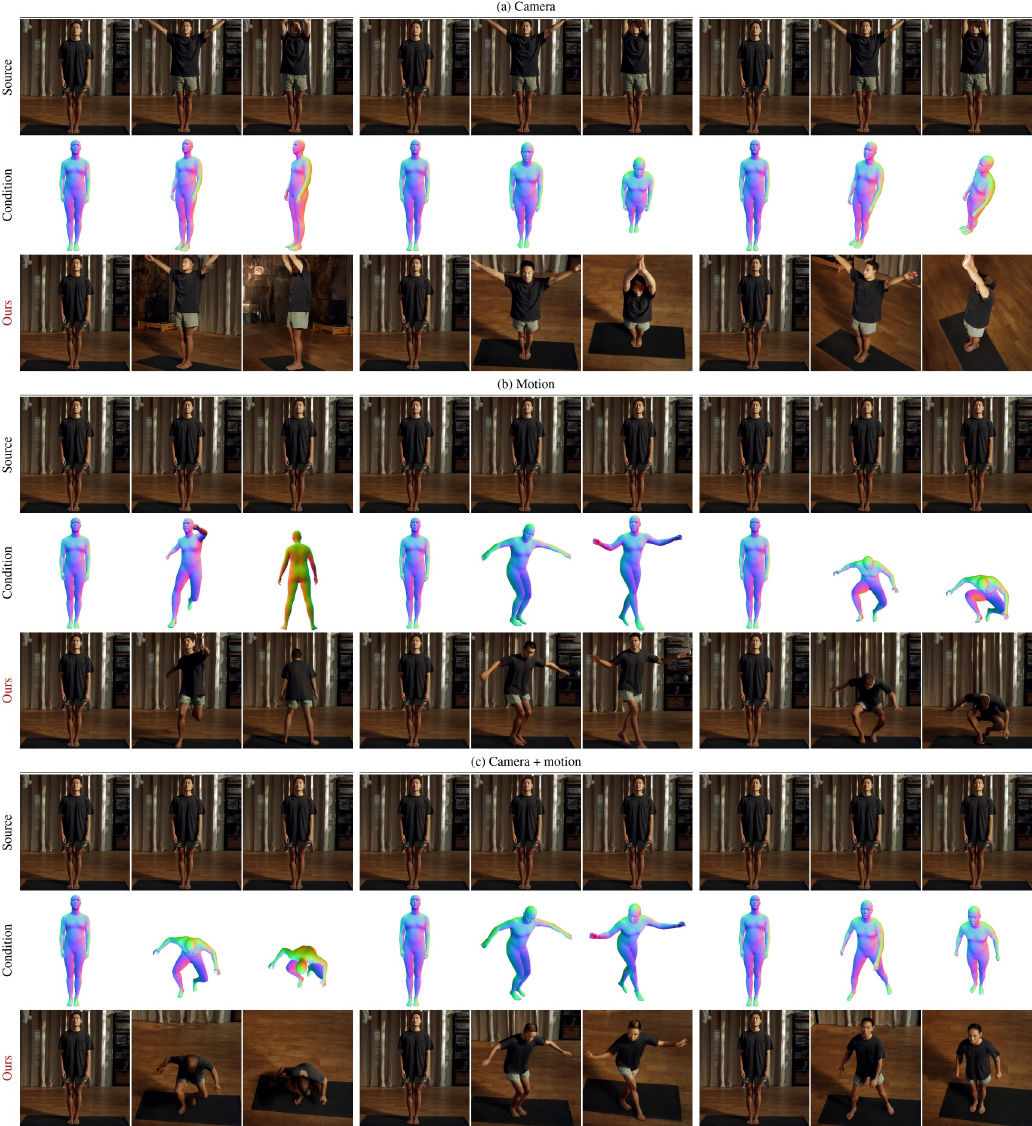}}
\caption{\textbf{Motion transferred from another video.} Actions extracted from an unrelated video drive a single reference image, under a fixed camera, a $45^\circ$ orbit, and a $45^\circ$ upward arc. Rows show the reference image (repeated), the retargeted condition, and \ourmethod{}; the driving video is not shown.}
\label{fig:supp:motion_transfer}
\end{figure*}

\vspace{1mm}
\noindent\textbf{Motion from another video.}
In the motion-control evaluation, each reference image is driven by its own video. Because the condition is a mesh sequence, motion can also come from an unrelated video. We fit SAM 3D Body to each driving frame and transfer the smoothed joint rotations, the root rotation relative to the first driving frame, and the lateral root displacement, while keeping the reference subject's body shape, scale, and hand pose. The driving videos in \cref{fig:supp:motion_transfer} are captured with static cameras; we do not estimate the driving camera's motion. The sequence starts from the reference pose and holds the final pose and position after the action ends. This retargeting is an image-based heuristic, not calibrated motion capture. \Cref{fig:supp:motion_transfer} transfers actions to a reference image, under a fixed camera, a $45^\circ$ orbit, and a $45^\circ$ upward arc; in the moving-camera cases, the camera orbits the center of the 3D bounding box of the reference subject's first-frame mesh and does not follow the subject's later displacement.

\section{More visual results}
\label{sec:supp:more_results}
\begin{figure*}[p]
\centering
\includegraphics[width=\linewidth,height=.9\textheight,keepaspectratio]{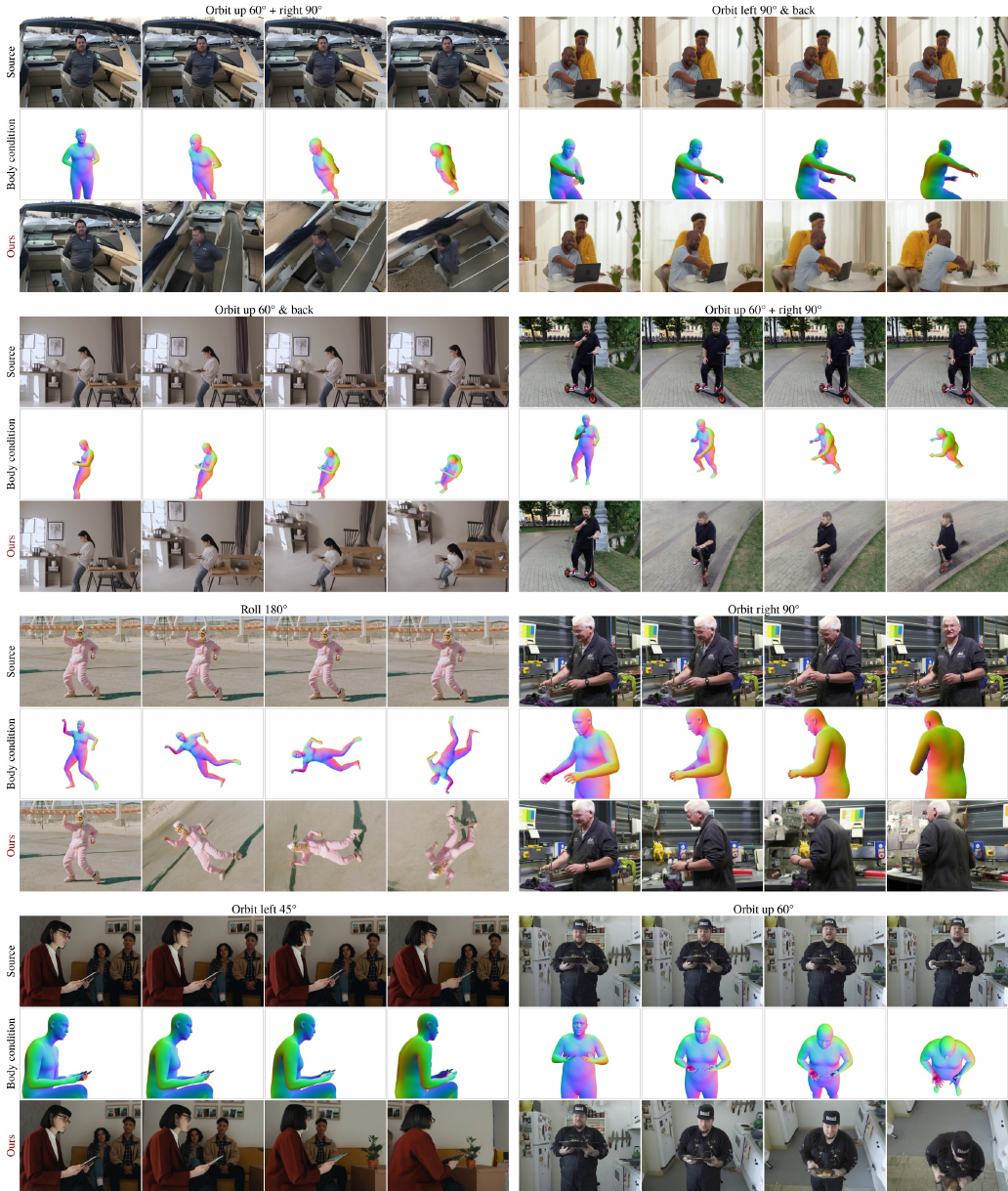}
\caption{\textbf{More visual results on in-the-wild scenes.}}
\label{fig:supp:more_scenes}
\end{figure*}

\begin{figure*}[p]
\centering
\includegraphics[width=\linewidth,height=.9\textheight,keepaspectratio]{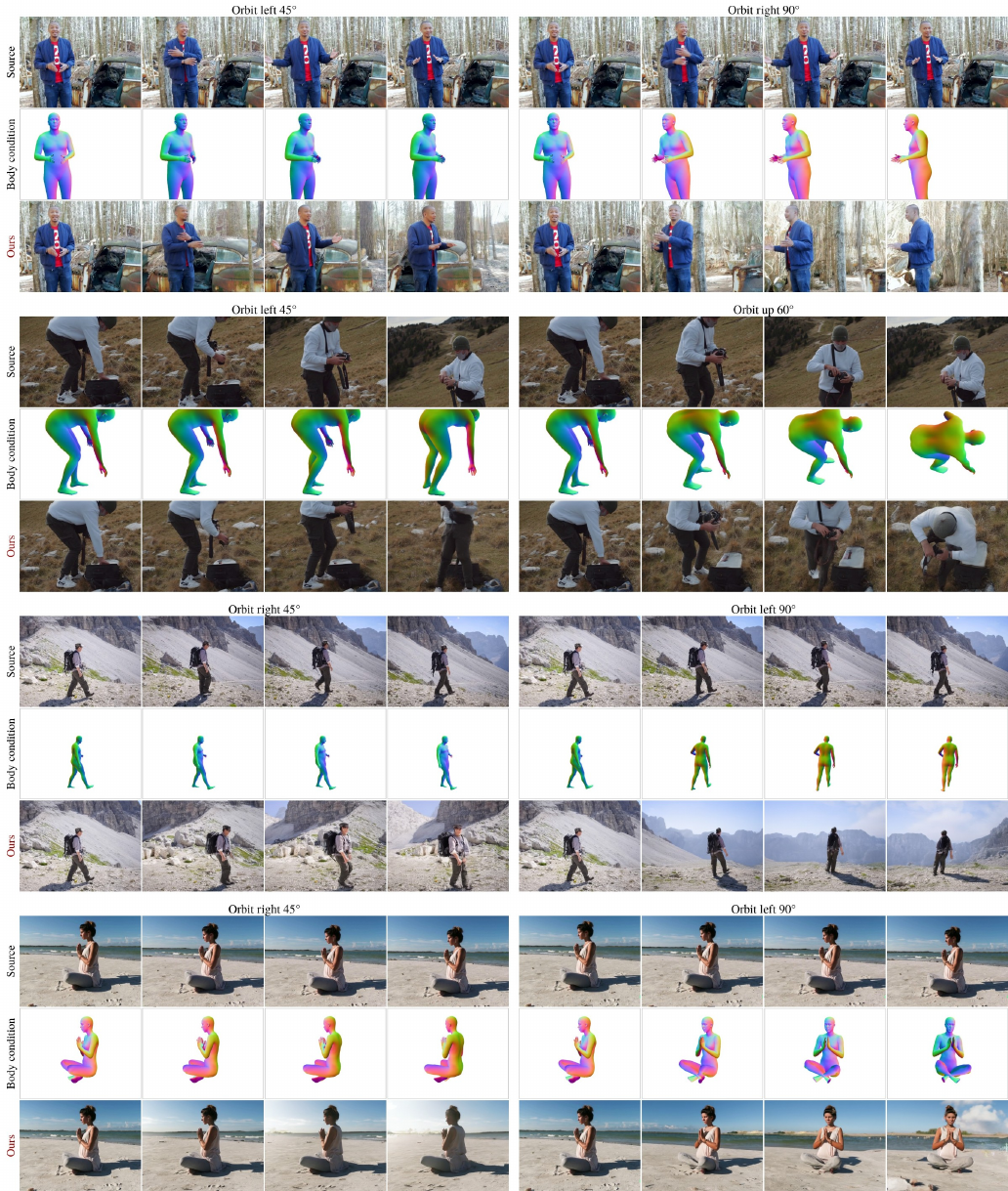}
\caption{\textbf{Different trajectories on the same scenes.}}
\label{fig:supp:more_orbits}
\end{figure*}

\Cref{fig:supp:more_scenes,fig:supp:more_orbits} show additional results of \ourmethod{} on in-the-wild videos.

\end{document}